\pdfoutput=1
\documentclass[nohyperref]{article}

\usepackage{natbib}
\usepackage[top=1in, bottom=1in, left=1in, right=1in]{geometry}
\usepackage[english]{babel}
\usepackage{etoc}
\usepackage{microtype}
\usepackage{graphicx}
\usepackage{xfrac}
\usepackage{subcaption}
\usepackage{booktabs}
\usepackage{multirow}
\usepackage{pifont}
\usepackage{wrapfig}
\usepackage{hyperref}
\usepackage{algorithm}
\usepackage{forloop}
\usepackage{tablefootnote}
\usepackage{algpseudocode}
\usepackage{enumitem}
\usepackage{amsmath}
\usepackage{amssymb}
\usepackage{mathtools}
\usepackage{amssymb}
\usepackage{pifont}
\usepackage{amsthm}
\usepackage[capitalize,noabbrev]{cleveref}
\usepackage{listings}

\usepackage{hyperref}
\usepackage{pgfplots}
\usepackage{multirow}
\usepackage{makecell}
\usepackage{amssymb}
\usepgfplotslibrary{fillbetween}
\usepackage{pifont}
\usepackage{dblfloatfix}

\usepackage{xcolor}
\definecolor{dkgreen}{rgb}{0,0.6,0}
\definecolor{dkred}{rgb}{0.8,0.0,0}
\definecolor{dkblue}{rgb}{0.0,0.0,0.9}
\definecolor{gray}{rgb}{0.5,0.5,0.5}
\definecolor{mauve}{rgb}{0.58,0,0.82}
\definecolor{lightgray}{HTML}{EEEEEE}
\definecolor{dkcyan}{HTML}{008b8b}

\usepackage{amsmath}
\usepackage{mathtools}
\usepackage{amsthm}
\usepackage{mathtools}
\usepackage{caption}
\usepackage{tikz}
\usepackage{latexsym}
\usepackage{algorithmicx}
\usepackage{algpseudocode}

\pgfplotsset{
    legend image with text/.style={
        legend image code/.code={%
            \node[anchor=center] at (0.3cm,0cm) {#1};
        }
    },
}

\usepgfplotslibrary{colorbrewer}

\usepackage[capitalize,noabbrev]{cleveref}

\theoremstyle{plain}

\theoremstyle{definition}

\theoremstyle{remark}

\usepackage{amsmath}
\usepackage{nicefrac}

\DeclareMathOperator*{\argmin}{arg\,min}

\DeclarePairedDelimiterX{\infdivx}[2]{(}{)}{%
  #1\;\delimsize\|\;#2%
}
\newcommand{\infdiv}{D_{KL}\infdivx}

\usepackage[textsize=tiny]{todonotes}

\usepackage[font=small,labelfont=bf]{caption}
\usepackage{tikz}
\usepackage{wrapfig}
\usepackage{thm-restate}

\hypersetup{
    colorlinks=true,
    linkcolor=dkcyan,
    citecolor=dkcyan,
    filecolor=dkcyan,
    urlcolor=dkcyan}
    
\everypar{\looseness=-1}
\setlist[itemize]{leftmargin=*}
\newif\ifcomments

\commentstrue

\ifcomments
  \newcommand{\colornote}[3]{{\color{#1}\bf{(#2) #3}\normalfont}}
\else
  \newcommand{\colornote}[3]{}
\fi

\newcommand{\E}{\mathbb{E}}

\newcommand{\reals}{\mathbb{R}}
\newcommand{\xmark}{\ding{55}}%

\newcommand{\is}{\mathcal{X}}
\newcommand{\os}{\mathcal{Y}}

\newcommand{\trainset}{\mathcal{D}_{\text{train}}}

\newcommand{\inp}{\mathbf{x}}
\newcommand{\target}{\mathbf{y}}
\newcommand{\htarget}{\hat{\target}}

\newcommand{\point}{\mathbf{z}}

\newcommand{\ucost}{\mathcal{Q}}

\newcommand{\param}{\boldsymbol{\theta}}
\newcommand{\cost}{\mathcal{J}}
\newcommand{\loss}{\mathcal{L}}

\newcommand{\fisher}{\mathbf{G}}

\newcommand{\x}{\mathbf{x}}

\newcommand{\oparam}{\param^{\star}}
\newcommand{\hparam}{\hat{\param}}

\title{Measuring Stochastic Data Complexity with Boltzmann Influence Functions}

\newcommand*\samethanks[1][\value{footnote}]{\footnotemark[#1]}

\author{Nathan Ng\thanks{Correspondence to: \texttt{nathanng@mit.edu}.}~\thanks{Massachusetts Institute of Technology.}~\thanks{University of Toronto and Vector Institute for Artificial Intelligence.},\,
Roger Grosse\samethanks[2],
Marzyeh Ghassemi\samethanks[3]
}
\date{}

\begin{document}
\etocdepthtag.toc{mtchapter}
\etocsettagdepth{mtchapter}{subsection}
\etocsettagdepth{mtappendix}{none}

\maketitle

\begin{abstract}
Estimating the uncertainty of a model's prediction on a test point is a crucial part of ensuring reliability and calibration under distribution shifts.
A minimum description length approach to this problem uses the predictive normalized maximum likelihood (pNML) distribution, which considers every possible label for a data point, and decreases confidence in a prediction
if other labels are also consistent with the model and training data.
In this work we propose IF-COMP, a scalable and efficient approximation of the pNML distribution that linearizes the model with a temperature-scaled Boltzmann influence function.
IF-COMP can be used to produce well-calibrated predictions on test points as well as measure complexity in both labelled and unlabelled settings.
We experimentally validate IF-COMP on uncertainty calibration, mislabel detection, and OOD detection tasks, where it consistently matches or beats strong baseline methods.

\end{abstract}

\section{Introduction}

Safely deploying machine learning models in real world settings requires making accurate predictions, as well as quantifying the uncertainty in those predictions.
This is particularly important in high-stakes settings such as healthcare \citep{Ghassemi2022}, medical imaging \citep{Esteva2017}, and self-driving cars \citep{bojarski2016end},
where uncertainty estimates can help accurately assess the risk in utilizing a model's decision and decide when to ignore it altogether \citep{ovadia2019trust}.
Common methods for quantifying uncertainty rely on Bayesian principles, which require defining a prior distribution and sampling from a posterior distribution.
However, specifying a good prior is difficult in the context of modern deep learning, and Bayesian methods face significant scalability challenges.

The Minimum Description Length (MDL) Principle \citep{rissanen1996fisher} provides an alternative approach to uncertainty estimation which does not require explicitly defining a prior or even a notion of ground truth.
MDL instantiates a version of Occam's Razor by favoring models that minimize the combined codelength of both the model and the observed data under a suitable coding scheme.
This code can then be used to make a prediction for an unseen test point by selecting the label with the shortest code.
However, this code may perform poorly compared to a hindsight-optimal code that is allowed to observe the test point and label.
We can attempt to minimize this difference in codelengths, or regret, by using a universal code such as the predictive normalized maximum likelihood (pNML) \citep{sequentially2008roos, 8437543} distribution, which minimizes the maximum possible regret for all labels.

Intuitively, given a query data point $\inp$, pNML considers augmenting the training set with the pair $(\inp, \target)$ for some label $\target$ and fitting a hindsight-optimal model to the augmented training set.
If one can just as easily fit the augmented dataset with arbitrary labels, then one is highly uncertain about the label; if one choice of labels is much easier to fit than the others, then one has low uncertainty.
This intuition is captured by the minimax optimality of pNML.
The universal codelength pNML defines for a data point is called the stochastic data complexity \citep{rissanen1986complexity}, and consists of two terms.
The error term captures the best possible loss a hindsight-optimal model could achieve by observing the true label.
The second parametric complexity term quantifies how many essentially different distributions the model class could assign to the test data.
Minimizing the pNML codelength then requires both learning an accurate model of the data as well as avoiding overfitting outlier observations.





Calculating the pNML distribution requires optimizing a hindsight-optimal model over the training set \textit{and the additional example for each possible label}, which is both computationally intractable and misspecified for overparameterized neural networks that can easily fit random labels \citep{zhang2017understanding}.
A simple way to restrict the hindsight-optimal model subclass is with a proximal objective that penalizes movement in function and weight space.
Approximating this proximal objective by linearizing the network and applying a second order approximation of the proximal terms has been shown to correspond to the influence function (IF) in neural networks \citep{bae2022influence}.

However, when models are overconfident on training examples, the proximal terms can become too restrictive, making the hindsight-optimal model unable to fit low probability labels.
In this work we propose to temperature-scale the proximal objective and approximate it with a Boltzmann influence function (BIF), allowing the hindsight-optimal model to better accommodate arbitrary labels.
We then propose \textbf{I}nfluence \textbf{F}unction \textbf{Comp}lexity \textbf{(IF-COMP)}, a complexity measure and associated pNML code which uses the BIF to produce estimates of the hindsight-optimal output probabilities. 
These estimates can then be used to produce calibrated output distributions as well as measure the stochastic complexity of both labelled and unlabelled examples.

We validate IF-COMP's ability to reliably estimate ground truth pNML complexity, then 
investigate its use on three tasks that test its different capabilities.
We consider (1) uncertainty calibration, which requires producing reliable uncertainty estimates under distribution shifts,
(2) mislabel detection, which requires estimating complexity for labelled training examples, and 
(3) OOD detection, which requires estimating complexity for unlabelled test examples.
Across all three tasks, IF-COMP displays strong performance, consistently matching or beating Bayesian and optimization tracing approaches that often use \textit{more} information than is available to our method, as well as a similar pNML approximation method that explicitly takes steps in parameter space \citep{zhou2021amortized}.
Compared to this baseline, IF-COMP also provides a 7-15 times speedup in computational efficiency.
IF-COMP demonstrates the potential of MDL-based approaches for uncertainty and complexity estimation in deep neural networks, enabling better calibrated decision making.







\section{Background and Preliminaries}

\subsection{Minimum Description Length and Stochastic Complexity}
\label{subsec:complexity}
In this work we consider a supervised classification task from an input space $\is$ to a discrete output space $\os$, where we are given a finite training set $\trainset = \{(\inp_i, \target_i)\}_n$ consisting of pairs of examples $(\inp_i, \target_i)$. 
We consider a hypothesis class of possible models $\{\param \in \boldsymbol{\Theta}\}$, each of which defines a conditional probability distribution $p_{\param}(\target | \inp) = \sigma(f_{\param}(\inp))$ where $\sigma$ is the softmax function.
We can define a \textit{prefix code} for $\target$ using the Kraft inequality \citep{elements1991cover}, such that the codelength $\loss(\param, \inp, \target)$, or number of bits required to describe $\target$ given $\inp$, is:
\begin{align}
\label{eq:log_loss}
    \loss(\param, \inp, \target) = -\log p_{\param}(\target | \inp),
\end{align}
or the log loss.



The Minimum Description Length (MDL) Principle \citep{mdl1978rissanen, grunwald2004tutorial} dictates that we should choose the model in our hypothesis class whose code can describe the training labels given the training inputs in the fewest number of bits. 
For parametric hypothesis classes such as neural networks, the MDL principle reduces to the maximum likelihood principle \citep{grunwald2004tutorial, JMLR:v24:21-1133}.
Then, a given learner $q$ will aim to find the parameters $\hparam \in \boldsymbol{\Theta}$ that minimizes the standard training objective:
\begin{align}
\label{eq:cost}
    \cost(\param, \trainset) = \sum_{i=1}^n \loss(\param, \inp_i, \target_i)
\end{align}
We call the resulting model the \textit{base model} and the codelength it defines for an example the \textit{base codelength}.

We now consider an unseen test example $\point = (\inp, \target)$, for which our base model may not achieve the optimal codelength. 
If $q$ is additionally given access to this test example, it 
will be able to find the parameters that minimize the total codelength on both the training set \textit{and} the test example:
\begin{align}
\label{eq:inf_obj}
    \oparam(\inp, \target, \epsilon) = \argmin_{\param \in \boldsymbol{\Theta}}\cost(\param, \trainset) -
    \epsilon \log p_{\param}(\target | \inp)
\end{align}
where $\epsilon$ represents the amount to weight the test point relative to the training set.
We call the resulting model the \textit{hindsight-optimal model} and the codelength it defines for an example the \textit{hindsight-optimal codelength}.
Typically we will always set $\epsilon = \nicefrac{1}{n}$ so that the test point is weighted equally to the other training points, but we include this additional parameter for reasons that will become clear later. 
For ease of notation, we define the hindsight-optimal parameters as $\oparam(\inp, \target) \coloneqq \oparam(\inp, \target, \nicefrac{1}{n})$.

The regret of the learner $q$ for a test point $\point$ is the difference between the base codelength and the hindsight-optimal codelength:
\begin{align}
\label{eq:regret}
    R(q, \trainset, (\inp, \target)) = -\log p_{\hparam}(\target | \inp) + \log p_{\oparam(\inp, \target)} (\target | \inp).
\end{align}
The predictive normalized maximum likelihood (pNML) distribution \citep{universal1987shtarkov,roos2008bayesian,sequentially2008roos} is then defined as the universal model which minimizes the worst case regret across all possible labels:
\begin{align}
\label{eq:pnml}
    p_{\text{pNML}}(\target | \inp) = \frac{p_{\oparam(\inp, \target)}(\target | \inp)}{\sum_{\target' \in \os} p_{\oparam(\inp, \target')}(\target' | \inp) }.
\end{align}
The codelength of the pNML distribution is
\begin{align}
\label{eq:pnml_log}
    \Gamma(\point) =  \overbrace{-\log p_{\oparam(\inp, \target)}(\target | \inp)}^{\text{error}} + \underbrace{\log \sum_{\target' \in \os} p_{\oparam(\inp, \target')}(\target' | \inp)}_{\text{parametric complexity}},
\end{align}
also known as the \textit{stochastic complexity} \citep{rissanen1996fisher, barron1998minimum} of $\point$ relative to the model class $\Theta$.
The first error term measures the hindsight-optimal codelength or log loss for $\point$. 
The second parametric complexity term quantifies how many distinguishable distributions the model class $\Theta$ could assign to the data in hindsight.
A model that is expressive enough to overfit its training data and any arbitrary test data  will be able to achieve low error but at the cost of high parametric complexity.
A less expressive model that underfits its training data and any arbitrary test data will have low parametric complexity but  will incur high error.
Choosing a model with a small description length under the pNML code necessitates trading off between these two extremes.

\subsection{The Infinity Problem and Proximal Bregman Objective}
\label{subsec:infinity}
For many overparameterized model classes, the denominator in \eqref{eq:pnml} is either infinite (for continuous label spaces) or constant (for discrete label spaces).
To solve this \textit{infinity problem}, the hindsight-optimal model must be restricted to a subclass of models that ``comply'' with the original training data \citep{fogel2019universal}.
This can be done by restricting the label space \citep{foster2001competitive}, enforcing moment matching \citep{farnia2017minimax}, confidence intervals, or minimum training likelihood \citep{fogel2019universal}.

Another approach directly alters the objective with a proximal term, such as in a ridge estimator \citep{JMLR:v24:21-1133}.
In this work we propose to use a
proximal bregman objective (PBO) \citep{bae2022influence} that restricts movement in function and weight space away from the base model optimum while training on the additional unseen test example:
\begin{align}
\label{eq:pbo}
    \ucost(\param, \trainset, \point) = -\log p_{\param}(\target | \inp) + \sum_{i=1}^n \infdiv{p_{\param}(\target_i | \inp_i)}{p_{\hparam}(\target_i | \inp_i)} + \frac{\lambda}{2}||\param - \hparam||_2^2.
\end{align}
Since the base model may not have reached convergence, the PBO penalizes movement in function space using the KL divergence between the base and hindsight model output distributions rather than using the true training labels.



\subsection{Influence Functions}
Influence functions \citep{75272a7e-1c8b-3ed5-9350-a7ee81abee59, bd831960-ac2b-396a-8c8f-de3944255f11} are a classical method from robust statistics that attempt to measure the sensitivity of an estimator to individual datapoints \citep{pmlr-v206-fisher23a}.
We can formulate this alternative objective using Eq. \ref{eq:inf_obj} which we rewrite as a response function $r_{(\inp, \target)} : \reals \to \Theta$
\begin{align*}
    r_{(\inp, \target)}(\epsilon) = \oparam(\inp, \target, \epsilon)
\end{align*}
where we assume that the objective (\ref{eq:inf_obj}) is strongly convex and hence the optimum is unique given some factor $\epsilon$.
Under these assumptions, note that $r_{(\inp, \target)}(0) = \hparam$ and the response function is differentiable at $\epsilon_0 = 0$ by the Implicit Function Theorem \citep{griewank2008evaluating, krantz2002implicit}. 
This allows us to approximate the response function with a first order Taylor expansion about $\epsilon_0$:
\begin{align*}
    r_{(\inp, \target)}(\epsilon) = \hparam + \fisher^{-1} \nabla_{\param} \loss(\hparam, \inp, \target) \epsilon,
\end{align*}
where $\fisher$ is the Hessian of \eqref{eq:cost} or a positive definite approximation such as the Generalized Gauss-Newton Hessian or Fisher Information. 

To approximate the effects on the loss at a specific test point $\point' = (\inp', \target')$, we can linearize the model about $\hparam$ and apply the chain rule:
\begin{align*}
    \loss(\oparam(\inp, \target, \epsilon), \inp', \target') = \loss(\hparam, \inp', \target') + \epsilon \nabla_{\param} \loss(\hparam, \inp', \target')^\intercal \fisher^{-1} \nabla_{\param} \loss(\hparam, \inp, \target)
\end{align*}
If the train and test point are the same, the quantity 
\begin{align}
    \text{IF}\ (\param, \inp, \target) = \nabla_{\param} \loss(\hparam, \inp, \target)^\intercal \fisher^{-1} \nabla_{\param} \loss(\hparam, \inp, \target)
\end{align}
is often referred to as the \textit{self-influence} of $\point$ relative to the model parameterized by $\param$ \citep{koh2017understanding}.

Although influence functions were meant to approximate the effects of true retraining \citep{basu2020influence}, recent work has shown that they more closely approximate an alternate objective \citep{bae2022influence, grosse2023studying}, which coincides with the proximal bregman objective \eqref{eq:pbo} described in Section \ref{subsec:infinity}.


\section{IF-COMP: Measuring Complexity with Boltzmann Influence Functions}
Although the pNML distribution and stochastic complexity are useful for calibrating the uncertainty of a given base model, exactly calculating them is intractable because they require finding the hindsight-optimal model for each possible label.
In addition, overparameterized neural networks can achieve arbitrarily low log loss on the additional test point regardless of its label, introducing the infinity problem.
We empirically verify this behavior in Appendix \ref{app:unrestricted_oracle}.
Methods such as ACNML \citep{zhou2021amortized} which take explicit steps in parameter space are also computationally expensive and can become unreliable for overconfident models.



To solve these issues, we begin by defining a temperature-scaled proximal Bregman objective that softens the local curvature, allowing the hindsight-optimal model to fit low probability labels.
Linearizing the model produces a corresponding Boltzmann influence function (BIF) which can directly approximate the hindsight-optimal output distribution.
We then propose \textbf{I}nfluence \textbf{F}unction \textbf{Comp}lexity \textbf{(IF-COMP)}, a complexity measure and associated pNML code that uses the BIF to produce calibrated output distributions as well as approximate the stochastic complexity for labelled and unlabelled data points.
Finally, we describe how to efficiently compute IF-COMP and then validate it against the ground truth pNML complexity.

\subsection{Boltzmann Influence Functions}
Since the PBO (Eq. \ref{eq:pbo}) regularizes movement in function space using the model's own output distribution, overconfident predictions can make this objective \textit{too} restrictive for training the hindsight-optimal model with very low probability labels.
Inspired by the use of 
temperature scaling to reduce model overconfidence and improve calibration for predictions \citep{pmlr-v70-guo17a}, as well as the alternative Boltzmannian MDL formulation \citep{perotti2018thermodynamics}, we propose to directly temperature scale our codelength as:
\begin{align}
\label{eq:alt_ham}
    E_{\beta, \param}(\inp, \target) = -\log p_{\beta, \param}(\target | \inp) \coloneqq -\log \sigma(\beta f_{\param}(\inp)).
\end{align}
which gives us a corresponding minimax optimal Boltzmann pNML distribution
\begin{align}
\label{eq:bnml}
    p_{\beta, \text{pNML}}(\target | \inp) = \frac{p_{\beta, \oparam(\inp, \target)}(\target | \inp)}{\sum_{\target' \in \os} p_{\beta, \oparam(\inp, \target')}(\target' | \inp) }.
\end{align}
and a stochastic complexity
\begin{align}
\label{eq:bnml_comp}
    \Gamma_{\beta}(\point) = {-\log p_{\beta, \oparam(\inp, \target)}(\target | \inp)} + \log \sum_{\target' \in \os} p_{\beta, \oparam(\inp, \target')}(\target' | \inp)
\end{align}

We can replace the loss in the original PBO with our temperature-scaled loss to define a corresponding hindsight-optimal model with a Boltzmann PBO (BPBO):
\begin{align}
\label{eq:bpbo}
    \ucost_{\beta}(\param, \trainset, \point) = E_{\beta, \param}(\inp, \target) + \sum_{i=1}^n \infdiv{p_{\beta, \param}(\target_i | \inp_i)}{p_{\beta, \hparam}(\target_i | \inp_i)} + \frac{\lambda}{2}||\param - \hparam||_2^2.
\end{align}
If we take a first order approximation of the log loss and a second order approximation of the proximal terms about $\hparam$, we can formulate the influence function as
\begin{align}
\label{eq:bif}
    \text{IF}_{\beta}(\hparam, \inp, \target) = \nabla_{\param} E_{\beta, \hparam}(\inp, \target)^\intercal \fisher_{\beta}^{-1} \nabla_{\param} E_{\beta, \hparam}(\inp, \target),
\end{align}
with a corresponding Fisher information:
\begin{align*}
    \fisher_{\beta} = \frac{1}{n} \sum_{i=1}^n \mathop{\mathbb{E}}_{\target_i' \sim p_{\beta, \hparam}(\target_i | \inp_i)} \left[ \nabla_{\param} E_{\beta, \hparam}(\inp_i, \target_i')\nabla_{\param} E_{\beta, \hparam}(\inp_i, \target_i')^\intercal \right].
\end{align*}
Intuitively, temperature scaling softens the function space distance loss and allows the model to better accommodate arbitrarily labelled data points.
We call the influence function calculated with this temperature-scaled loss the \textbf{Boltzmann influence function (BIF)}.

Different choices for $\beta$ define different BIFs. At $\beta = 1$, we recover the standard influence function.
As $\beta \to 0^+$, the KL divergence loss becomes an MSE loss over logits \citep{kim2021comparing} and we recover a BIF resembling TRAK \citep{park2023trak}.
In Section \ref{subsec:ablations} we will see that tuning $\beta$ directly within the influence function is an important part of achieving high quality complexity estimates.

\subsection{IF-COMP}
We now describe \textbf{IF-COMP}, which estimates the Boltzmann pNML distribution and stochastic complexity using our Boltzmann influence function.
Recall from \eqref{eq:bnml} and \eqref{eq:bnml_comp} that our goal is to approximate $p_{\beta, \oparam(\inp, \target)}(\target | \inp)$ for all $\target \in \os$.
Using the BIF formulation above, we can apply a first order Taylor expansion of the output probability about $\hparam$ to approximate
\begin{align*}
    p_{\beta, \oparam(\inp, \target)}(\target | \inp) = p_{\beta, \hparam}(\target | \inp) + \frac{1}{n} p_{\beta, \hparam}(\target | \inp)\ \text{IF}_{\beta}(\hparam, \inp, \target).
\end{align*}
Similar to other work \citep{ilyas2022datamodels, park2023trak}, we choose to linearize our model in the log probability space, where we have used the identity
\begin{align*}
    \nabla p_{\beta, \hparam}(\target | \inp) = p_{\beta, \hparam}(\target | \inp) \nabla \log  p_{\beta, \hparam}(\target | \inp)
\end{align*}
to change the log probability gradient into a probability space gradient.

We can then rewrite the Boltzmann pNML parametric complexity as
\begin{align*}
    \log \sum_{\target' \in \os} p_{\beta, \oparam(\inp, \target')}(\target' | \inp) = \log \left( 1 + \frac{1}{n} \E_{\target' \sim p_{\beta, \hparam}(\target | \inp)} \left[\text{IF}_{\beta}(\hparam, \inp, \target') \right] \right).
\end{align*}
For large enough $n$ we can further simplify $\log(1 + x) \approx x$, which gives us a final complexity
\begin{align}
\label{eq:ifcomp_full}
    \Gamma_{\beta}(\point) = \overbrace{-\log p_{\beta, \oparam(\inp, \target)}(\target | \inp)}^{\text{error}} + \underbrace{\frac{1}{n} \E_{\target' \sim p_{\beta, \hparam}(\target | \inp)}\left[ \text{IF}_{\beta}(\hparam, \inp, \target') \right].}_{\text{parametric complexity}}
\end{align}
We leave the error term in its original form, since we assume access to true labels only for training data, for which our model is already hindsight-optimal.
For unlabelled data that may not have a valid label, the error term is undefined and we compute only the parametric complexity term.
\begin{align}
\label{eq:ifcomp_par}
    \Gamma_{\beta}(\inp) = \E_{\target' \sim p_{\beta, \hparam}(\target | \inp)}\left[ \text{IF}_{\beta}(\hparam, \inp, \target') \right]
\end{align}

If we are interested in the calibrated probabilities of the Boltzmann pNML distribution, we can calculate the distribution directly as
\begin{align}
\label{eq:pnml_dist}
    p_{\beta, \text{pNML}}(\target | \x) = \frac{p_{\beta, \hparam}(\target | \inp) + \nicefrac{\alpha}{n}\ p_{\beta, \hparam}(\target | \inp)\ \text{IF}_{\beta}(\hparam, \inp, \target)}{1 + \nicefrac{\alpha}{n}\ \E_{\target' \sim p_{\beta, \hparam}(\target | \inp)} \left[\text{IF}_{\beta}(\hparam, \inp, \target') \right]}
\end{align}
where $\alpha$ controls the weighting of the test point relative to the training set.
A value of $\alpha=0$ corresponds to the original model output distribution.
Compared to ACNML \citep{zhou2021amortized}, which explicitly optimizes the hindsight-optimal parameters using a weighted approximate posterior, IF-COMP linearizes the model directly which allows us to easily control this weighting \textit{after} computing the BIF for each label.
    

\subsection{Efficiently Computing IF-COMP}

\begin{table*}[t]
    \centering
    \small
    \caption{Average pNML output distribution \eqref{eq:pnml_dist} inference time per input (in seconds). On smaller models such as LeNet, IF-COMP can be computed almost as efficiently as a gradient norm while ACNML is nearly 7 times slower. On larger ResNet18 models, IF-COMP maintains a greater than 10 times speedup compared to ACNML.}
    \vspace{0.2cm}
    \resizebox{0.95\textwidth}{!}{%
        \begin{tabular}{llll}
\toprule
Method & \textbf{MNIST LeNet} & \textbf{CIFAR-10 ResNet18} & \textbf{CIFAR-100 ResNet18 }\\
\midrule
GradNorm & 0.000226s & 0.00336s & 0.03783s \\
ACNML-EKFAC & 0.001770s & 0.23183s & 2.33628s \\
IF-COMP (ours) & 0.000241s & 0.01576s & 0.18883s \\
\midrule
IF-COMP vs. ACNML Speedup & 7.34 $\times$ & 14.71 $\times$ & 12.4 $\times$ \\
\bottomrule
\end{tabular}
    }
    \label{tab:timing}
\end{table*}

Although we have simplified the calculation of stochastic data complexity considerably, calculating the BIF still requires estimating and inverting the Fisher information matrix. 
Similar to previous work \citep{grosse2023studying}, we use an eigenvalue-corrected Kronecker-factored approximation to the Fisher information matrix (EKFAC) \citep{george2021fast} which produces an eigendecomposition $\fisher_{\beta} \approx Q\lambda Q^T$. 
We can then calculate the influence function as the squared L2 norm in the inverse eigenspace:
\begin{align}
    \text{IF}_{\beta}(\hparam, \inp, \target) = || \nabla_{\param} E_{\beta, \param}(\inp, \target)^\intercal Q (\lambda + \delta)^{-\nicefrac{1}{2}} ||_2^2,
\end{align}
where $\delta$ is a damping term added to ensure invertibility.
Since we need only calculate and invert the EKFAC once for a given model and training set, calculating the IF-COMP then requires only one jacobian vector product (JVP) per label per sample.

We verify the computational efficiency of this approach by comparing the time to calculate the final pNML output distribution \eqref{eq:pnml_dist} for a single example with IF-COMP, ACNML, and gradient norm methods. 
Specific experimental details are provided in Appendix \ref{app:experimental_details}.
We present our results in Table \ref{tab:timing}.
For small LeNet models, we find that the additional overhead of the EKFAC multiplication performed in IF-COMP is relatively small (6\%) compared to a simple gradient norm, while ACNML is nearly 7 times slower.
For larger ResNet18 models, IF-COMP becomes slower than the gradient norm, but still provides a 12-15 times speedup over ACNML.

\begin{figure}[t]
    \centering
    \resizebox{0.6\columnwidth}{!}{%
        \begin{tikzpicture}
\begin{axis}[
    ybar,
    symbolic x coords={CIFAR-10, CIFAR-100, MNIST, All},
    xtick=data, 
    bar width=10, 
    enlarge x limits=0.18,
    ylabel=Pearson R,
    ymin=0,
    ymax=1.0,
    ymajorgrids=true,
    height=6cm, width=11cm,
    xtick pos=left,
    ytick pos=left,
    legend style={at={(0.5,1.2)},anchor=north,column sep=0.2cm},
    legend cell align={left},
    legend style={legend columns=4},
    legend image code/.code={
        \draw [#1] (0cm,-0.1cm) rectangle (0.2cm,0.25cm); },
    ]
    ]
    
    \addplot[fill=blue!60, fill opacity=0.8,] coordinates {
        (CIFAR-10, 0.569)
        (CIFAR-100, 0.796)
        (MNIST, 0.19)
        (All, 0.631)
    };
    
    \addplot[fill=red!60, fill opacity=0.8,] coordinates {
        (CIFAR-10, 0.346)
        (CIFAR-100, 0.743)
        (MNIST, 0.120)
        (All, 0.455)
    };
    
    \addplot[fill=yellow!80, fill opacity=0.8,] coordinates {
        (CIFAR-10, 0.836)
        (CIFAR-100, 0.818)
        (MNIST, 0.344)
        (All, 0.839)
    };


    \addplot[fill=black!30!green, fill opacity=0.8] coordinates {
        (CIFAR-10, 0.928)
        (CIFAR-100, 0.8882)
        (MNIST, 0.406)
        (All, 0.8707)
    };
    
    \legend{GradNorm, Self-IF, ACNML-EKFAC, IF-COMP}

\end{axis}
\end{tikzpicture}
    }
    \caption{Pearson R correlation of different methods of approximating hindsight-optimal outputs with ground truth parametric complexity on in-domain (CIFAR-10) and out-of-domain datasets. IF-COMP achieves the highest correlation across all datasets, beating ACNML, a computationally more expensive alternative.}
    \label{fig:oracle_val}
\end{figure}
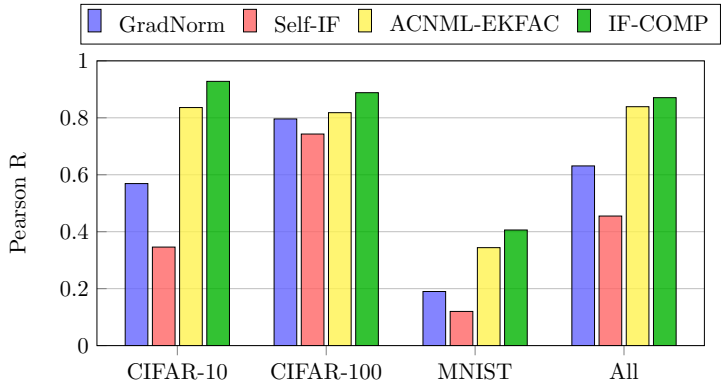

\subsection{pNML Validation}
\label{subsec:oracle_val}
To verify that IF-COMP can accurately approximate the ground truth pNML parametric complexity on both in-distribution (ID) and out-of-distribution (OOD) samples, we fine-tune
a CIFAR-10 \citep{Krizhevsky2009learning} pre-trained ResNet-18 \citep{he2016resnet} model with the BPBO (\ref{eq:bpbo}) on 20 random test images each from CIFAR-10, CIFAR-100, and MNIST \citep{deng2012mnist}. 
We label each example with every possible output class for a total of 600 individual training example pairs.
After training, we can use the hindsight-optimal output probabilities to calculate the complexity as in Eq. \ref{eq:ifcomp_par}.

We compare IF-COMP with other baselines for approximating hindsight-optimal training. 
Specifically, we consider averaging gradient norms across classes, self influence calculated with EKFAC, and ACNML with an EKFAC posterior. 
We present our results in Figure \ref{fig:oracle_val}.
We find that IF-COMP consistently achieves the highest Pearson correlation with true complexity on all datasets, beating ACNML which explicitly takes steps in parameter space.
For ID CIFAR-10 examples, IF-COMP achieves a strong Pearson R of 0.928, which it maintains on CIFAR-100 examples.
All methods degrade considerably on MNIST, which indicates that complexity becomes more difficult to estimate for further OOD examples.

\section{Experiments}

To experimentally validate IF-COMP, we consider three tasks that test different capabilities.
The first, uncertainty quantification, requires producing calibrated outputs across distribution shifts.
The second, mislabel detection, requires measuring complexity on labelled training examples.
We use this task as a case study to understand how the error and parametric complexity terms trade off during training, as well as the effects of temperature scaling.
Finally, OOD detection requires measuring complexity on unlabelled out-of-distribution test examples.
Across all three tasks, IF-COMP consistently exhibits strong performance compared to baseline methods.
We provide additional experiments on data pruning and hindsight-optimal retraining in Appendix \ref{app:extra experiments} as well as full experimental details in Appendix \ref{app:experimental_details}.


\begin{figure}
    \centering
    \begin{subfigure}[t]{0.32\textwidth}
         \centering
         \resizebox{\textwidth}{!}{%
             \begin{tikzpicture}
\begin{axis}[
    xlabel=Confidence,
    ylabel=Accuracy,
    xtick pos=left,
    ytick pos=left,
    ymajorgrids=true,
    xmin=0.3, xmax=1,
    ymin=0.3, ymax=1,
    height=7cm, width=8cm,
    legend pos=south east,
    legend style={nodes={scale=0.8, transform shape}},
    title=CIFAR-10C Pixelate Level 1
]

\addplot [black, dotted, forget plot, thick]  {x};

\addplot+[red, mark options={red}, mark size=1.2pt,  mark=*, error bars/.cd, y dir=both, y explicit,] table [x=x, y=y, col sep=comma] {
x,  y
0.5024452805519104, 0.496
0.739465594291687, 0.656
0.8858099579811096, 0.828
0.9479743838310242, 0.91
0.9701435565948486, 0.954
0.9795498251914978, 0.972
0.9842259287834167, 0.994
0.9868590831756592, 0.994
0.9886232614517212, 0.988
0.9899592995643616, 0.994
0.9910444617271423, 0.994
0.9919534921646118, 0.996
0.9927447438240051, 0.996
0.9934417009353638, 0.996
0.9941386580467224, 0.998
0.9947934150695801, 1.0
0.9954931139945984, 0.998
0.9962482452392578, 0.996
0.9970769286155701, 0.998
0.9983670115470886, 1.0
};
\addlegendentry{SWA + Temp}

\addplot+[orange, mark options={orange}, mark size=1.2pt,  mark=*, error bars/.cd, y dir=both, y explicit,] table [x=x, y=y, col sep=comma] {
x,  y
0.46610355377197266, 0.472
0.6662368774414062, 0.646
0.8106646537780762, 0.754
0.9092094302177429, 0.868
0.9621589183807373, 0.948
0.983908474445343, 0.932
0.992878794670105, 0.974
0.9963887333869934, 0.99
0.9979416728019714, 0.988
0.998647153377533, 0.988
0.9990418553352356, 0.998
0.9992789626121521, 0.994
0.9994261264801025, 0.998
0.9995367527008057, 1.0
0.9996228218078613, 0.996
0.999692976474762, 1.0
0.9997546672821045, 1.0
0.9998094439506531, 1.0
0.9998660087585449, 1.0
0.9999310374259949, 1.0
};
\addlegendentry{SWAGD}

\addplot+[blue, mark options={blue}, mark size=1.2pt,  mark=*, error bars/.cd, y dir=both, y explicit,] table [x=x, y=y, col sep=comma] {
x,  y
0.42042890191078186, 0.47
0.5398436784744263, 0.66
0.680404782295227, 0.784
0.8418894410133362, 0.85
0.947942852973938, 0.916
0.9833580255508423, 0.954
0.9937726855278015, 0.976
0.9969081878662109, 0.984
0.9982129335403442, 0.992
0.9988493919372559, 0.99
0.9991578459739685, 0.998
0.9993464350700378, 0.998
0.9994758367538452, 0.99
0.9995732307434082, 0.994
0.9996508955955505, 0.996
0.9997183680534363, 1.0
0.9997795224189758, 0.998
0.9998345971107483, 1.0
0.9998872876167297, 1.0
0.9999464154243469, 1.0
};
\addlegendentry{ACNML + SWAGD}

\addplot+[purple!60!black, mark options={purple!60!black}, mark size=1.2pt,  mark=*, error bars/.cd, y dir=both, y explicit,] table [x=x, y=y, col sep=comma] {
x,  y
0.48948559165000916, 0.48
0.6891394257545471, 0.74
0.8355490565299988, 0.866
0.9218644499778748, 0.918
0.9660996198654175, 0.96
0.985666036605835, 0.982
0.9934399127960205, 0.98
0.9965226054191589, 0.996
0.9978374242782593, 0.988
0.9984903931617737, 0.996
0.9988275170326233, 1.0
0.9990243315696716, 1.0
0.9991487264633179, 1.0
0.9992461800575256, 1.0
0.9993250966072083, 1.0
0.9993953704833984, 1.0
0.9994593858718872, 0.998
0.9995235204696655, 1.0
0.999596118927002, 0.998
0.9997142553329468, 1.0
};
\addlegendentry{Ensemble}

\addplot+[black!60!green, mark options={black!60!green}, mark size=1.2pt,  mark=*, error bars/.cd, y dir=both, y explicit,] table [x=x, y=y, col sep=comma] {
x,  y
0.4299602806568146, 0.524
0.5586177706718445, 0.614
0.7363007068634033, 0.84
0.875036358833313, 0.902
0.9355190992355347, 0.958
0.9621126055717468, 0.978
0.9740251898765564, 0.992
0.9800118207931519, 0.988
0.9835213422775269, 0.994
0.9858382344245911, 0.988
0.987562358379364, 0.996
0.9888783097267151, 0.998
0.9900496602058411, 0.998
0.9910144805908203, 0.998
0.9919239282608032, 0.998
0.9928650856018066, 1.0
0.9938008785247803, 0.998
0.9947590231895447, 0.998
0.9958503246307373, 1.0
0.9975370168685913, 1.0
};
\addlegendentry{IF-COMP}


\end{axis}
\end{tikzpicture}
         }   
         \label{fig:pixel_1}
    \end{subfigure}
    \hfil
    \begin{subfigure}[t]{0.32\textwidth}
         \centering
         \resizebox{\textwidth}{!}{%
             \begin{tikzpicture}
\begin{axis}[
    xlabel=Confidence,
    ylabel=Accuracy,
    xtick pos=left,
    ytick pos=left,
    ymajorgrids=true,
    xmin=0.3, xmax=1,
    ymin=0.3, ymax=1,
    height=7cm, width=8cm,
    legend pos=north west,
    title=CIFAR-10C Pixelate Level 3
]

\addplot [black, dotted, forget plot, thick]  {x};

\addplot+[red, mark options={red}, mark size=1.2pt,  mark=*, error bars/.cd, y dir=both, y explicit,] table [x=x, y=y, col sep=comma] {
x,  y
0.40499624609947205, 0.356
0.5513781309127808, 0.482
0.6768459677696228, 0.564
0.7918968796730042, 0.666
0.8721573948860168, 0.756
0.9205808043479919, 0.828
0.9504736065864563, 0.894
0.9669567942619324, 0.92
0.9766988754272461, 0.934
0.9821020364761353, 0.958
0.9853372573852539, 0.982
0.9876253604888916, 0.982
0.9893274307250977, 0.984
0.9906837940216064, 0.99
0.9917930364608765, 0.99
0.9928455948829651, 0.994
0.9938479065895081, 0.998
0.994835376739502, 0.994
0.9960271120071411, 1.0
0.997599184513092, 0.996
};

\addplot+[orange, mark options={orange}, mark size=1.2pt,  mark=*, error bars/.cd, y dir=both, y explicit,] table [x=x, y=y, col sep=comma] {
x,  y
0.36648809909820557, 0.34
0.5130467414855957, 0.466
0.6064366102218628, 0.518
0.7032670378684998, 0.608
0.799943208694458, 0.694
0.8732801675796509, 0.782
0.9255053997039795, 0.834
0.9589904546737671, 0.894
0.978229820728302, 0.876
0.9894489049911499, 0.946
0.9945929050445557, 0.976
0.9970651865005493, 0.966
0.9982385039329529, 0.98
0.9988587498664856, 0.984
0.9992233514785767, 0.994
0.9994349479675293, 0.998
0.9995778799057007, 0.996
0.9996860027313232, 0.994
0.9997758269309998, 1.0
0.9998835325241089, 0.998
};

\addplot+[blue, mark options={blue}, mark size=1.2pt,  mark=*, error bars/.cd, y dir=both, y explicit,] table [x=x, y=y, col sep=comma] {
x,  y
0.34944865107536316, 0.398
0.45049938559532166, 0.518
0.5005137324333191, 0.566
0.5520859360694885, 0.62
0.6373851299285889, 0.686
0.7459685206413269, 0.758
0.8580560088157654, 0.792
0.9333840608596802, 0.866
0.9733428955078125, 0.912
0.9893567562103271, 0.94
0.9950615167617798, 0.958
0.9974997043609619, 0.98
0.9985060691833496, 0.982
0.9990088939666748, 0.994
0.9993005990982056, 0.994
0.9994750022888184, 0.988
0.9996067881584167, 0.99
0.9997079968452454, 1.0
0.9997991919517517, 0.998
0.9999042749404907, 0.996
};

\addplot+[purple!60!black, mark options={purple!60!black}, mark size=1.2pt,  mark=*, error bars/.cd, y dir=both, y explicit,] table [x=x, y=y, col sep=comma] {
x,  y
0.3954029679298401, 0.376
0.5145097374916077, 0.514
0.5995813608169556, 0.568
0.6940327286720276, 0.646
0.7795923948287964, 0.736
0.8562350869178772, 0.8
0.912055253982544, 0.84
0.9495647549629211, 0.912
0.9727981686592102, 0.942
0.9873613715171814, 0.952
0.993739128112793, 0.976
0.9965759515762329, 0.97
0.9978856444358826, 0.99
0.9985681176185608, 0.988
0.9989434480667114, 0.996
0.9991501569747925, 1.0
0.9992846846580505, 1.0
0.999401867389679, 0.998
0.9995054006576538, 0.996
0.9996501207351685, 0.998
};

\addplot+[black!60!green, mark options={black!60!green}, mark size=1.2pt,  mark=*, error bars/.cd, y dir=both, y explicit,] table [x=x, y=y, col sep=comma] {
x,  y
0.36549079418182373, 0.384
0.4617668390274048, 0.5
0.517313539981842, 0.568
0.6063138246536255, 0.646
0.7199098467826843, 0.75
0.8158654570579529, 0.822
0.8872939944267273, 0.894
0.9315595030784607, 0.9
0.9562331438064575, 0.946
0.9691855311393738, 0.956
0.9768825769424438, 0.98
0.9814608097076416, 0.984
0.9845107197761536, 0.99
0.9867819547653198, 0.988
0.9885794520378113, 0.996
0.9901792407035828, 0.998
0.9915785789489746, 0.998
0.9929706454277039, 1.0
0.9944843053817749, 1.0
0.9965369701385498, 0.996
};


\end{axis}
\end{tikzpicture}
         }   
         \label{fig:pixel_3}
    \end{subfigure}
    \hfil
    \centering
    \begin{subfigure}[t]{0.32\textwidth}
         \centering
         \resizebox{\textwidth}{!}{%
             \begin{tikzpicture}
\begin{axis}[
    xlabel=Confidence,
    ylabel=Accuracy,
    xtick pos=left,
    ytick pos=left,
    ymajorgrids=true,
    xmin=0.3, xmax=1,
    ymin=0.3, ymax=1,
    height=7cm, width=8cm,
    legend pos=north west,
    title=CIFAR-10C Pixelate Level 5
]

\addplot [black, dotted, forget plot, thick]  {x};

\addplot+[red, mark options={red}, mark size=1.2pt,  mark=*, error bars/.cd, y dir=both, y explicit,] table [x=x, y=y, col sep=comma] {
x,  y
0.32702577114105225, 0.24
0.424443781375885, 0.228
0.48751384019851685, 0.28
0.5433191061019897, 0.28
0.6005286574363708, 0.316
0.6588841676712036, 0.332
0.7152678370475769, 0.362
0.7681304812431335, 0.386
0.8159760236740112, 0.388
0.8605046272277832, 0.474
0.896268367767334, 0.458
0.9248901605606079, 0.522
0.9475399851799011, 0.592
0.9641045928001404, 0.626
0.975534200668335, 0.682
0.9822112917900085, 0.804
0.9866542220115662, 0.884
0.9897539019584656, 0.884
0.9922768473625183, 0.926
0.9954715371131897, 0.95
};

\addplot+[orange, mark options={orange}, mark size=1.2pt,  mark=*, error bars/.cd, y dir=both, y explicit,] table [x=x, y=y, col sep=comma] {
x,  y
0.29943159222602844, 0.188
0.4011845886707306, 0.196
0.46305760741233826, 0.266
0.5128355622291565, 0.312
0.5591838359832764, 0.322
0.6082372665405273, 0.324
0.6553249955177307, 0.348
0.7092350721359253, 0.406
0.7600477337837219, 0.39
0.8059905767440796, 0.428
0.847775399684906, 0.414
0.8899285793304443, 0.478
0.928695797920227, 0.514
0.9572716951370239, 0.572
0.9779353737831116, 0.65
0.9900326728820801, 0.712
0.9958404898643494, 0.78
0.9982286691665649, 0.872
0.9992199540138245, 0.93
0.9996837973594666, 0.966
};

\addplot+[blue, mark options={blue}, mark size=1.2pt,  mark=*, error bars/.cd, y dir=both, y explicit,] table [x=x, y=y, col sep=comma] {
x,  y
0.303400456905365, 0.174
0.37515392899513245, 0.238
0.41452062129974365, 0.274
0.4478136897087097, 0.304
0.47557613253593445, 0.406
0.4953252077102661, 0.358
0.5152871608734131, 0.38
0.548831045627594, 0.38
0.5922089219093323, 0.352
0.6490011215209961, 0.424
0.7167425751686096, 0.448
0.7942002415657043, 0.474
0.8706132769584656, 0.478
0.929542601108551, 0.536
0.969832718372345, 0.618
0.9881378412246704, 0.69
0.9956115484237671, 0.75
0.99836665391922, 0.864
0.9992454648017883, 0.928
0.9997015595436096, 0.954
};

\addplot+[purple!60!black, mark options={purple!60!black}, mark size=1.2pt,  mark=*, error bars/.cd, y dir=both, y explicit,] table [x=x, y=y, col sep=comma] {
x,  y
0.3397669196128845, 0.212
0.4193879961967468, 0.248
0.47266510128974915, 0.234
0.514387309551239, 0.288
0.5557452440261841, 0.306
0.5971909761428833, 0.292
0.6412802934646606, 0.35
0.6867192387580872, 0.336
0.7333102226257324, 0.336
0.7802472114562988, 0.358
0.8265994191169739, 0.378
0.8715805411338806, 0.46
0.9121034145355225, 0.464
0.9467679262161255, 0.542
0.9721266031265259, 0.642
0.9883420467376709, 0.702
0.9958328604698181, 0.79
0.9982513189315796, 0.88
0.9990466237068176, 0.918
0.9994584918022156, 0.956
};

\addplot+[black!60!green, mark options={black!60!green}, mark size=1.2pt,  mark=*, error bars/.cd, y dir=both, y explicit,] table [x=x, y=y, col sep=comma] {
x,  y
0.31100979447364807, 0.238
0.3783496022224426, 0.242
0.41748979687690735, 0.242
0.4491174817085266, 0.334
0.4753127694129944, 0.334
0.5002444982528687, 0.372
0.5349793434143066, 0.36
0.5791112184524536, 0.396
0.632114052772522, 0.412
0.6935557723045349, 0.432
0.7597413659095764, 0.47
0.8216474652290344, 0.524
0.8766410946846008, 0.578
0.9199410676956177, 0.61
0.9501492977142334, 0.68
0.9683119058609009, 0.79
0.9787721037864685, 0.874
0.9849480986595154, 0.896
0.9892259240150452, 0.934
0.9936376810073853, 0.96
};


\end{axis}
\end{tikzpicture}
         }   
         \label{fig:pixel_5}
    \end{subfigure}
    \caption{Reliability diagrams for Pixelate corruptions on CIFAR-10C. IF-COMP outperform ACNML as well as Bayesian methods and ensembles even as corruptions increase in severity. Although IF-COMP and ACNML perform similarly on lower confidence examples, IF-COMP maintains this reliability on higher confidence examples. Dotted lines represent perfect calibration.}
    \label{fig:pixel_rel}
\end{figure}
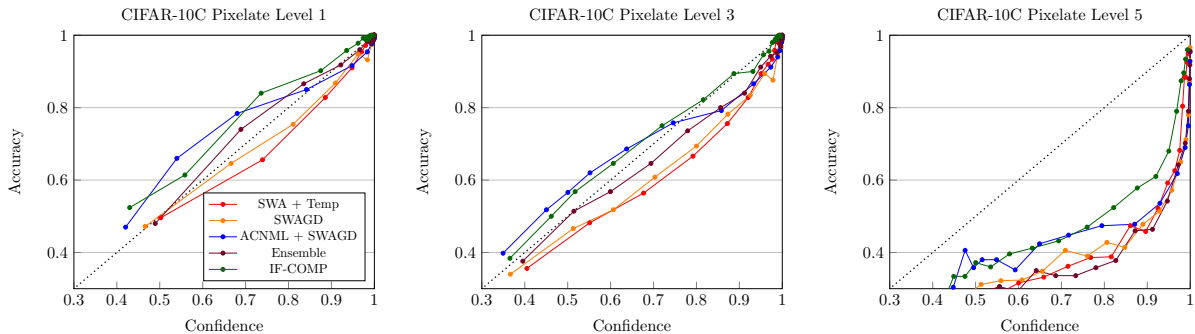

\begin{figure}[t]
    \centering
    \resizebox{0.6\columnwidth}{!}{%
        \pgfplotsset{
}

\begin{tikzpicture}
\begin{axis}[
    label style={align=center,font=\large},
    xlabel=Corruption Severity,
    ylabel=ECE,
    xtick pos=left,
    xtick={1,2,3,4,5},
    ytick pos=left,
    ymajorgrids=true,
    xmin=0.5,xmax=5.5,
    ymin=0,
    height=8cm, width=12cm,
    legend pos=north west,
    yticklabel style={
        /pgf/number format/fixed,
        /pgf/number format/precision=5
    },
    legend style={column sep=10pt,},
    scaled y ticks=false,
]

\addplot+[brown, mark options={brown}, mark=*, dashed, error bars/.cd, error bar style={solid}, y dir=both, y explicit] table [x=x, y=y,y error plus=plus, y error minus=minus, col sep=comma] {
x,  y, minus, plus
0.75, 0.020170442366600035, 0.0063277703523635775, 0.009528764867782598
1.75, 0.038393574583530425, 0.008207273340225224, 0.03162076050043106
2.75, 0.06930505409240723, 0.030367981195449825, 0.05461029332876208
3.75, 0.11139600435495375, 0.05680936644673344, 0.08969386318325998
4.75, 0.18520607227087021, 0.085615226238966, 0.15058309109807017
};
\addlegendentry{MC-Dropout}

\addplot+[orange, mark options={orange}, mark=*, dashed, error bars/.cd, y dir=both, y explicit, error bar style={solid}] table [x=x, y=y,y error plus=plus, y error minus=minus, col sep=comma] {
x,  y, minus, plus
0.85, 0.018537040793895728, 0.006514969718456273, 0.01967475535869598
1.85, 0.0376443190097809, 0.01431529916524888, 0.03956628888249395
2.85, 0.057913749015331265, 0.026183627182245248, 0.054204592150449774
3.85, 0.07982194123268128, 0.021082663148641587, 0.11683922894597053
4.85, 0.144939632999897, 0.05165716550350191, 0.12363237380385403
};
\addlegendentry{SWAGD}

\addplot+[red, mark options={red}, mark=*, dashed, error bars/.cd, error bar style={solid}, y dir=both, y explicit] table [x=x, y=y,y error plus=plus, y error minus=minus, col sep=comma] {
x,  y, minus, plus
0.95, 0.01738920110464096, 0.004345582008361819, 0.01978683091998101
1.95, 0.030144973313808446, 0.010724674338102345, 0.04657674470543863
2.95, 0.049820598709583296, 0.024298571783304225, 0.06791075589656828
3.95, 0.06924911174774169, 0.022818784618377677, 0.13679497259855272
4.95, 0.1510931219816208, 0.06630253638029102, 0.13008216189742086
};
\addlegendentry{SWA + Temp}

\addplot+[blue, mark options={blue}, mark=*, dashed, error bars/.cd, error bar style={solid}, y dir=both, y explicit] table [x=x, y=y,y error plus=plus, y error minus=minus, col sep=comma] {
x,  y, minus, plus
1.05, 0.023000424742698675, 0.007728983247280122, 0.002528202778100966
2.05, 0.026629821145534523, 0.006447905552387237, 0.012741412073373797
3.05, 0.033134620475769046, 0.007199035018682475, 0.029135266304016118
4.05, 0.039716506612300875, 0.006434993666410441, 0.08517852650284768
5.05, 0.08513050109148024, 0.03822904753088949, 0.12144173532724381
};
\addlegendentry{ACNML + SWAGD}

\addplot+[purple, mark options={purple}, mark=*, dashed, error bars/.cd, error bar style={solid}, y dir=both, y explicit] table [x=x, y=y,y error plus=plus, y error minus=minus, col sep=comma] {
x,  y, minus, plus
1.15, 0.007513292431831365, 0.0026752216219902095, 0.0034762313544750203
2.15, 0.012165180361270914, 0.005756448131799706, 0.02159702743291854
3.15, 0.021413151752948764, 0.010377042853832246, 0.04883461998701097
4.15, 0.04293897697925567, 0.01729518415927886, 0.11045593144893648
5.15, 0.09884070206880573, 0.0437734599888325, 0.1584150098264217
};
\addlegendentry{Ensemble}

\addplot+[black!60!green, mark options={black!60!green}, mark=*, solid, error bars/.cd, error bar style={solid}, y dir=both, y explicit] table [x=x, y=y,y error plus=plus, y error minus=minus, col sep=comma] {
x,  y, minus, plus
1.25, 0.018880169999599454, 0.0011697775125503566, 0.003129634749889372
2.25, 0.018065447652339934, 0.0019906707167625416, 0.007656888210773458
3.25, 0.018468521428108217, 0.0022592041790485407, 0.017104476451873782
4.25, 0.019424964034557346, 0.0031334698140621092, 0.09020848699212075
5.25, 0.05493680689334869, 0.03095899371504783, 0.13733831776976585
};
\addlegendentry{IF-COMP}



\end{axis}
\end{tikzpicture}
    }    
    \caption{Expected calibration error (ECE) for various methods across increasing levels of CIFAR-10C corruptions. We plot medians and inter-quartile ranges. IF-COMP achieves lower ECE across almost all corruption levels compared to both Bayesian methods and other NML-based methods.}
    \label{fig:cifar10_ece}
\end{figure}
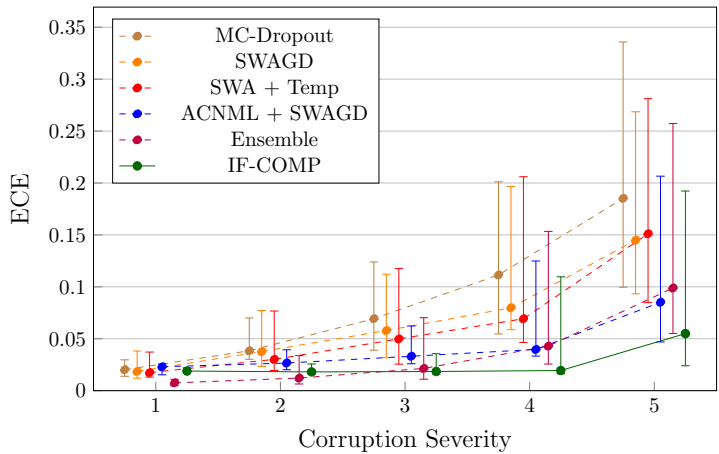

\subsection{Uncertainty Calibration}
We begin by evaluating the uncertainty calibration of IF-COMP output distributions under distribution shifts. 
Following the experimental setup in \citet{ovadia2019trust}, we measure the expected calibration error (ECE) \citep{naeini2015ece} of ResNet18 models trained on CIFAR-10 and tested on the CIFAR10-C datasets \citep{hendrycks2019robustness}.
CIFAR-10C applies 19 corruptions with 5 severity levels across the test images of CIFAR-10, allowing us to compare calibration on a wide range of distribution shifts.
We calculate ECE by dividing model predictions sorted by confidence into 20 equal sized bins \citep{zhou2021amortized}.

We compare against a wide range of Bayesian and NML baselines, including 
ensembling \citep{balaji2017ensembles}, Stochastic Weight Averaging (SWA) \citep{izmailov2019averaging}, SWA Gaussian Diagonal (SWAG-D) \citep{maddox2019swag}, Monte-Carlo dropout \citep{gal2016dropout}, and ACNML \citep{zhou2021amortized} with a SWAG-D posterior.
For a fair comparison, all methods except ensembling and MC-Dropout use the same SWA base model.
Additionally, we apply the same IF-COMP temperature scaling to the SWA output logits, which is equivalent to using a value of $\alpha = 0$.
We produce 30 model samples for all Bayesian methods.

We present median ECE and inter-quartile ranges across corruptions for each severity level in Figure \ref{fig:cifar10_ece}. 
On more severe corruptions (3-5), IF-COMP produces better calibrated uncertainties than all baselines, beating the most relevant ACNML baseline as well as a Bayesian ensemble of 30 trained models.
At corruption severity 2 IF-COMP outperforms all other baselines except the ensemble.
At the lowest corruption level all methods perform similarly.

The reliability diagrams in Figure \ref{fig:pixel_rel} show a more detailed look at the calibration for varying Pixelate corruption severity levels.
For low corruption levels all methods perform similarly. 
As severity increases, IF-COMP maintains strong calibration compared to Bayesian baselines and improves over ACNML for high confidence outputs.
Although all methods degrade considerably at the highest corruption severity, IF-COMP still improves over the strong ACNML baseline.

\begin{table*}[t]
    \centering
    \small
    \caption{Mislabel detection AUROC for CIFAR-10 and CIFAR-100 with various types and rates of label noise (best method bolded). IF-COMP achieves strong detection across all CIFAR-10 noise types without requiring extra checkpoints, even on the difficult data-dependent noise that other methods fail to perform better than random on. On CIFAR-100, IF-COMP achieves strong performance on symmetric and asymmetric noise, although it fails to detect mislabelled data with human and data-dependent noise, similar to other methods.}
    \vspace{0.2cm}
    \resizebox{0.95\textwidth}{!}{%
        \begin{tabular}{lcccccccccc}
\toprule
& \multirow{2.9}{*}{\makecell{Extra\\Checkpoints?}} & \multicolumn{4}{c}{CIFAR10} & \multicolumn{4}{c}{CIFAR100} \\
\cmidrule(lr){3-6} \cmidrule(lr){7-10}
Method & & Human & Data & Asym 0.3 & Sym 0.6 & Human & Data & Asym 0.3 & Sym 0.6\\
\midrule
Trac-IN & \checkmark & 90.50 & 49.13 & 70.98 & 64.40 & 50.01 & 50.25 & 58.25 & 57.95 \\
EL2N & \checkmark & 73.38 & 52.18 & 51.53 & 51.61 & 49.79 & \textbf{50.51} & 56.30 & 55.45 \\
GraNd & \checkmark & 70.53 & 51.30 & 50.48 & 50.52 & \textbf{50.52} & 50.07 & 52.26 & 50.16 \\
Self-IF & \xmark & 95.38 & 55.77 & 87.30 & 43.16 & 49.81 & 50.29 & 72.15 & 58.40 \\
\midrule
IF-COMP & \xmark & \textbf{96.86} & \textbf{88.07} & \textbf{95.69} & \textbf{97.83} & 49.52 & 50.30 & \textbf{79.39} & \textbf{95.21}  \\
\bottomrule
\end{tabular}
    }
    
    \label{tab:mislabel_auroc}
\end{table*}

\subsection{Mislabel Detection}
Next, we investigate IF-COMP's ability to measure the complexity of labelled training examples.
We follow the mislabel detection setup in \citet{clusterability2021zhu,srikanth2023empirical} and use CIFAR-10 and CIFAR-100 datasets corrupted with various types of label noise and noise rates.
We train a ResNet-18 model on this corrupted dataset then attempt to identify which examples were mislabelled using only the trained model and noised dataset.
We consider 4 types of label noise.
Human noise replaces all labels with labels from a single human annotator \citep{clusterability2021zhu, wei2021learning}.
Data-dependent noise is generated by jointly modelling the clean label and a class-dependent projection of the feature vector \citep{clusterability2021zhu}.
Symmetric noise changes the label to another uniformly at random.
Asymmetric noise changes each label to another fixed similar label. 

For each example, we calculate the full stochastic complexity (\ref{eq:ifcomp_full}) by combining the model's log error on the true label as well as the parametric complexity across all labels.
For our baselines, we consider the self-influence score (Self-IF) \citep{koh2017understanding}, Trac-IN \citep{tracin2020pruthi}, EL2N \citep{el2n2021paul}, and GraNd \citep{el2n2021paul} and measure performance using AUROC.
Trac-IN requires evaluating intermediate checkpoints, and EL2N and GraNd require multiple short training runs, meaning they utilize \textit{more} information than is available to IF-COMP.

We present our results in Table \ref{tab:mislabel_auroc}.
For CIFAR-10 datasets, 
IF-COMP consistently achieves the highest AUROC across all types of noise, beating even the baselines that utilize additional checkpoint information. 
On the difficult data-dependent noise where all other methods achieve close to random AUROC, IF-COMP maintains strong performance.
For CIFAR-100 datasets where all baselines consistently achieve close to random AUROC across all noise types, IF-COMP is still able to detect both asymmetric and symmetric noise.
On human and data-dependent noise no method performs better than random.

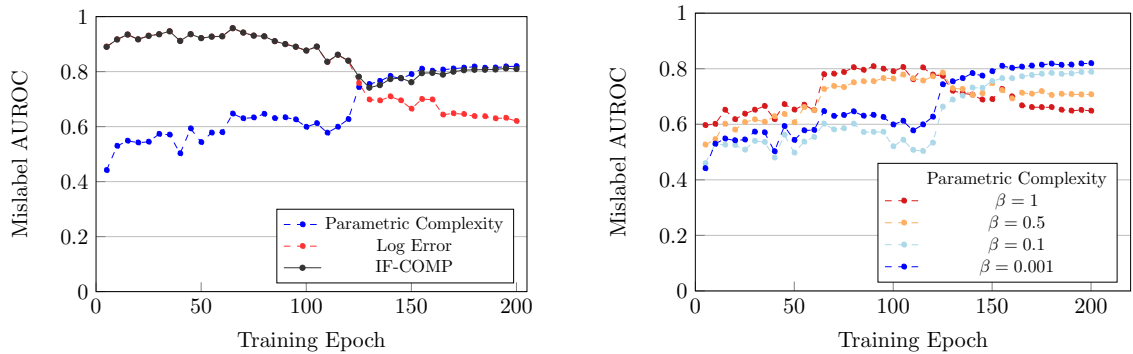
\begin{figure*}
    \vspace{-0.2cm}
    \centering
    \subfloat[AUROC of the individual components of IF-COMP.]{\centering\resizebox{0.45\columnwidth}{!}{%
        \begin{tikzpicture}
\begin{axis}[
    xlabel=Training Epoch,
    ylabel=Mislabel AUROC,
    xtick pos=left,
    ytick pos=left,
    ymajorgrids=true,
    xmin=0, xmax=205,
    ymin=0, ymax=1,
    height=6cm, width=8.5cm,
    legend pos=south east,
    legend style={nodes={scale=0.8, transform shape}},
]


\addplot+[blue, dashed, mark options={blue}, mark size=1.2pt,  mark=*, error bars/.cd, y dir=both, y explicit,] table [x=x, y=y, col sep=comma] {
x,  y
5, 0.44128989017597947
10, 0.529601470181603
15, 0.5482209572188074
20, 0.5417609171331323
25, 0.5446082441870548
30, 0.5730290500664699
35, 0.5703577578572832
40, 0.5024660675090096
45, 0.5929097911205203
50, 0.5431747712417596
55, 0.5776759373446471
60, 0.5791140305191227
65, 0.6468975027001683
70, 0.6297121055528084
75, 0.6328931919913066
80, 0.6459953698204943
85, 0.6303359492178288
90, 0.6333824259722676
95, 0.625714589734114
100, 0.5987559226333985
105, 0.6120149308136599
110, 0.5774684809808532
115, 0.5990009730083834
120, 0.6269139604043648
125, 0.7434311529839653
130, 0.754213687857152
135, 0.7656549565137722
140, 0.7836641086866064
145, 0.7746250758662806
150, 0.7904784607886084
155, 0.8100435973843513
160, 0.8027526777061476
165, 0.8069890576978149
170, 0.8108014307422421
175, 0.8139040083210313
180, 0.818005367332273
185, 0.8135127299013016
190, 0.8144186771739422
195, 0.8183653857048512
200, 0.8195235353392891
};
\addlegendentry{Parametric Complexity}

\addplot+[red!80, dashed, mark options={red!80}, mark size=1.2pt,  mark=*, error bars/.cd, y dir=both, y explicit,] table [x=x, y=y, col sep=comma] {
x,  y
5, 0.8902660320040792
10, 0.9168176406144815
15, 0.9346943980103541
20, 0.9172103600948128
25, 0.9301662093401438
30, 0.9357603548026473
35, 0.9464271981012999
40, 0.9115275315191562
45, 0.9359846819340548
50, 0.921789263130296
55, 0.9271096512547734
60, 0.9282811101349576
65, 0.9577427578318894
70, 0.9417303200347903
75, 0.9306373462809916
80, 0.9279104087200972
85, 0.9102085619588927
90, 0.8995269500740533
95, 0.8899610161422216
100, 0.8758802012626061
105, 0.890322335429973
110, 0.8351001772166362
115, 0.8606369211833473
120, 0.8387143328323103
125, 0.7580365763786752
130, 0.6980848207112373
135, 0.6947304002741415
140, 0.7094636804759056
145, 0.6948970575617278
150, 0.664751670785617
155, 0.6999290858038589
160, 0.6982034583864093
165, 0.6432306326002256
170, 0.647888336972834
175, 0.6451925391168207
180, 0.6377528779160787
185, 0.6371878892377423
190, 0.6296117248663364
195, 0.6315442054129095
200, 0.6200045104892167
};
\addlegendentry{Log Error}

\addplot+[black!80, mark options={black!80}, mark size=1.2pt,  mark=*, error bars/.cd, y dir=both, y explicit,] table [x=x, y=y, col sep=comma] {
x,  y
5, 0.8902655521583187
10, 0.9168140699526949
15, 0.9346899483929997
20, 0.9172102427991824
25, 0.930167339643491
30, 0.9357627266116924
35, 0.9464314969099867
40, 0.9115274416432836
45, 0.9359888345040336
50, 0.9218009043407795
55, 0.927121059397316
60, 0.9282839115202075
65, 0.9578240604602489
70, 0.9418949954832195
75, 0.9308447889349066
80, 0.9282978102907453
85, 0.9108057505727432
90, 0.9000689229124008
95, 0.8905126117922184
100, 0.876673302522239
105, 0.891029427002833
110, 0.8359506284229169
115, 0.8615122695296622
120, 0.839797589968422
125, 0.7814889769938514
130, 0.7417968394530927
135, 0.7510462991109682
140, 0.7729914493526715
145, 0.7767154820309173
150, 0.7614641420662008
155, 0.7944653179381387
160, 0.7956066973440744
165, 0.7892713607933128
170, 0.7998892948607444
175, 0.8052838564765671
180, 0.8067940971269545
185, 0.8070488571428215
190, 0.8078779666376557
195, 0.8118803601265892
200, 0.8097107062921042
};
\addlegendentry{IF-COMP}


\end{axis}
\end{tikzpicture}
    }\label{subfig:mislabel_ifcomp}}
    \hfil
    \subfloat[Parametric complexity AUROC at varying temperatures.]{\centering\resizebox{0.45\columnwidth}{!}{%
        \pgfplotsset{
    cycle list/RdYlBu-4,
}

\begin{tikzpicture}
\begin{axis}[
    xlabel=Training Epoch,
    ylabel=Mislabel AUROC,
    xtick pos=left,
    ytick pos=left,
    ymajorgrids=true,
    xmin=0,
    ymin=0, ymax=1,
    height=6cm, width=8.5cm,
    legend pos=south east,
    legend style={nodes={scale=0.8, transform shape}},
]


\addlegendimage{legend image with text=}
\addlegendentry{Parametric Complexity}

\addplot+[dashed, mark size=1.2pt,  mark=*, error bars/.cd, y dir=both, y explicit,] table [x=x, y=y, col sep=comma] {
x,  y
5, 0.5964885782953405
10, 0.6009358649703183
15, 0.6516451606347053
20, 0.6176669471982082
25, 0.6375824397485632
30, 0.6519165568269254
35, 0.6654352803826065
40, 0.6179526214676151
45, 0.6719084073093262
50, 0.6523975572639629
55, 0.6697346740242163
60, 0.6509136426276524
65, 0.7798119175839711
70, 0.7817055275772979
75, 0.7875506430358199
80, 0.8048070497391451
85, 0.7954721196497323
90, 0.8084878445060635
95, 0.799610842650864
100, 0.7903602405030824
105, 0.8055530286217699
110, 0.7615447500606976
115, 0.8041384372263232
120, 0.7783724244785304
125, 0.7736023189023178
130, 0.720277872982731
135, 0.7122208224232782
140, 0.706236789332091
145, 0.6884804503723239
150, 0.6900719540447542
155, 0.7267763281490096
160, 0.6999016294864349
165, 0.6674573488943384
170, 0.6613501325414726
175, 0.6611563936731517
180, 0.6613039576694267
185, 0.651903089155912
190, 0.6486986349905338
195, 0.6512315899519274
200, 0.6484855712862807
};
\addlegendentry{$\beta = 1$}

\addplot+[dashed, mark size=1.2pt,  mark=*, error bars/.cd, y dir=both, y explicit,] table [x=x, y=y, col sep=comma] {
x,  y
5, 0.5263805931046542
10, 0.5456461490438529
15, 0.6014010746735332
20, 0.5801205473698997
25, 0.6072729122092995
30, 0.6173740325894362
35, 0.6086325574568608
40, 0.6283078681625509
45, 0.6367260613046496
50, 0.6064675954475902
55, 0.6614517806300804
60, 0.650251680929404
65, 0.7266652994566125
70, 0.7373487621750998
75, 0.7336066370267007
80, 0.7504832252457148
85, 0.7543490439679502
90, 0.7546148967991457
95, 0.7656502921083157
100, 0.7642328749545011
105, 0.7782485039300328
110, 0.765059638536728
115, 0.7569488513453707
120, 0.7720555414247326
125, 0.7854650673185062
130, 0.7291865964510041
135, 0.7267968548843231
140, 0.7051874671928311
145, 0.7112895773774763
150, 0.7477214691624411
155, 0.7223122438257721
160, 0.6926322602145003
165, 0.7130381383756416
170, 0.7097159193973798
175, 0.7188692834675914
180, 0.7049916185264835
185, 0.7092618238359716
190, 0.7081616349994513
195, 0.7067860619192787
200, 0.7074973060621175
};
\addlegendentry{$\beta = 0.5$}

\addplot+[dashed, mark size=1.2pt,  mark=*, error bars/.cd, y dir=both, y explicit,] table [x=x, y=y, col sep=comma] {
x,  y
5, 0.4589785739301455
10, 0.5278565544878981
15, 0.5261370371064666
20, 0.5248210121235211
25, 0.5080588754966628
30, 0.5395961394455665
35, 0.5358009692524861
40, 0.4798007266668426
45, 0.5617342029162187
50, 0.49755063264677823
55, 0.5370755584115162
60, 0.553941501671245
65, 0.6018119963030488
70, 0.5810316861166738
75, 0.5850903373319047
80, 0.6016677622840202
85, 0.5716232070970644
90, 0.5720268624907772
95, 0.5723666755485355
100, 0.5202045942894802
105, 0.543538803562208
110, 0.5074673079331504
115, 0.5034707959822535
120, 0.5334413467663701
125, 0.6630113052483784
130, 0.688537236690618
135, 0.7025999910897974
140, 0.7316042025848355
145, 0.7306226773199249
150, 0.7549296878046402
155, 0.7655126144580282
160, 0.7653048625701788
165, 0.7720730702665324
170, 0.7767881154458692
175, 0.7813314291591167
180, 0.7846294075482102
185, 0.7810471380641562
190, 0.7825950991612842
195, 0.7877286612430325
200, 0.7883743234186201
};
\addlegendentry{$\beta = 0.1$}

\addplot+[blue, dashed, mark options={blue}, dashed, mark size=1.2pt,  mark=*, error bars/.cd, y dir=both, y explicit,] table [x=x, y=y, col sep=comma] {
x,  y
5, 0.44128989017597947
10, 0.529601470181603
15, 0.5482209572188074
20, 0.5417609171331323
25, 0.5446082441870548
30, 0.5730290500664699
35, 0.5703577578572832
40, 0.5024660675090096
45, 0.5929097911205203
50, 0.5431747712417596
55, 0.5776759373446471
60, 0.5791140305191227
65, 0.6468975027001683
70, 0.6297121055528084
75, 0.6328931919913066
80, 0.6459953698204943
85, 0.6303359492178288
90, 0.6333824259722676
95, 0.625714589734114
100, 0.5987559226333985
105, 0.6120149308136599
110, 0.5774684809808532
115, 0.5990009730083834
120, 0.6269139604043648
125, 0.7434311529839653
130, 0.754213687857152
135, 0.7656549565137722
140, 0.7836641086866064
145, 0.7746250758662806
150, 0.7904784607886084
155, 0.8100435973843513
160, 0.8027526777061476
165, 0.8069890576978149
170, 0.8108014307422421
175, 0.8139040083210313
180, 0.818005367332273
185, 0.8135127299013016
190, 0.8144186771739422
195, 0.8183653857048512
200, 0.8195235353392891
};
\addlegendentry{$\beta = 0.001$}


\end{axis}
\end{tikzpicture}
    }\label{subfig:mislabel_parcomp}}
    \caption{IF-COMP accurately trades off between log error and parametric complexity, maintaining strong AUROC throughout training. Tuning the temperature is critical to achieving accurate complexity estimates near convergence.}
    \label{fig:mislabel_trace}
\end{figure*}

\subsection{Analyzing the Components of IF-COMP}
\label{subsec:ablations}
We take a brief detour and use mislabel detection as a case study to understand how each component of IF-COMP contributes to the final complexity estimate.
We consider CIFAR-10 human mislabel noise and train a model to convergence, taking checkpoints every 5 epochs. 
For each model, we calculate IF-COMP, log error, and temperature-scaled parametric complexity. 
We plot the mislabel detection AUROC across training for these values in Figure \ref{fig:mislabel_trace}.

Figure \ref{subfig:mislabel_ifcomp} shows that mislabelled examples are ignored early in training, increasing their error and making them easy to distinguish.
Later in training the model learns to memorize these examples and decrease their error, increasing parametric complexity accordingly.
Choosing only one of these values to detect mislabelled examples with would fail in different scenarios, while IF-COMP accurately captures the tradeoff between the two throughout all stages of training.
Figure \ref{subfig:mislabel_parcomp} shows that temperature-scaling is a crucial component of IF-COMP. 
With a standard temperature $\beta=1$, parametric complexity estimates are inaccurate and mirror log error, providing no additional signal for mislabel detection.
As we increase the temperature we soften the second order restrictions, making complexity estimates more accurate and providing a complementary signal to the error.


\begin{table}
    \centering
    \caption{AUROC for OOD detection methods on near and far distribution shifts. Best methods are bolded, and second best methods are starred. We show only the most common baseline methods, although we compare against all methods in the OpenOOD benchmark. IF-COMP achieves a new state of the art on MNIST and CIFAR-10 benchmarks, beating out all 20 other baselines.}
    \vspace{0.2cm}
    \resizebox{0.6\columnwidth}{!}{%
        \begin{tabular}{lcccccc}
\toprule
& \multicolumn{2}{c}{MNIST} & \multicolumn{2}{c}{CIFAR-10} & \multicolumn{2}{c}{CIFAR-100} \\
\cmidrule(lr){2-3} \cmidrule(lr){4-5} \cmidrule(lr){6-7}
Method & Near & Far & Near & Far & Near & Far \\
\midrule
MSP & 91.45 & 98.51 & 88.03 & 90.73 & 80.27 & 77.76 \\
ODIN & 92.38 & 99.02 & 82.87 & 87.96 & 79.90 & 79.28 \\
Energy & 90.77 & 98.77 & 87.58 & 91.21 & 80.91$^{*}$ & 79.77 \\
MLS & 92.49 & 99.08 & 87.52 & 91.10 & \textbf{81.05} & 79.67 \\
KNN & 96.52 & 96.66 & 90.64$^{*}$ & 92.96 & 80.18 & 82.40$^{*}$ \\
GradNorm & 76.55 & 96.39 & 54.90 & 57.55 & 70.13 & 69.14 \\
RMDS & 98.00$^{*}$ & 98.12 & 89.80 & 92.20 & 80.15 & \textbf{82.92} \\
VIM & 94.63 & 98.99 & 88.68 & 93.48$^{*}$ & 74.98 & 81.70 \\
Gram & 73.90 & 99.75$^{*}$ & 58.66 & 71.73 & 51.66 & 73.36 \\
\midrule 
IF-COMP & \textbf{99.40} & \textbf{99.97} & \textbf{92.23} & \textbf{95.63} & 79.40 & 79.41\\
\bottomrule
\end{tabular}
    }
    \label{tab:ood}
    \vspace{-0.5cm}
\end{table}

\subsection{OOD Detection}
Finally, we investigate IF-COMP's ability to measure complexity on unlabelled data by detecting OOD examples.
We use the OpenOOD benchmark \citep{yang2022openood, zhang2023openood} and consider the ID MNIST, CIFAR-10, and CIFAR-100 datasets using the provided pretrained LeNet \citep{lecun1998lenet} and ResNet-18 models.
For each dataset, the benchmark provides a set of Near-OOD datasets which exhibit similar visual features to the ID training set, and a set of Far-OOD datasets that do not.
Given a test example from one of these datasets, an OOD detection method aims to assign a high score to OOD examples and a low score to ID examples.
We use the IF-COMP parametric complexity (\ref{eq:ifcomp_par}) as our score function, and measure performance using AUROC.
We compare our method against the full suite of 20 baseline approaches implemented in OpenOOD, although we show only 9 of the most common or most similar methods in our results.
We refer readers to \citet{zhang2023openood} for more details on baseline methods.

Our results are shown in Table \ref{tab:ood}.
On MNIST, IF-COMP achieves a new state of the art AUROC, beating all 20 baselines with near perfect detection on both Near and Far OOD datasets.
IF-COMP similarly achieves a new state of the art AUROC for CIFAR-10, beating the next best method by over 2 AUROC on Far OOD datasets.
On CIFAR-100 IF-COMP ranks in the middle of the baselines.
For this dataset all baselines perform quite similarly, with only small AUROC gaps between the best and second best method.
Strong baselines exhibit variance in performance across datasets, with the best method, RMDS \citep{ren2021simple}, beating all other baselines only on 2 datasets.
In contrast, IF-COMP achieves the best AUROC across 4 of the 6 datasets.

\section{Related Work}

\paragraph{The Minimum Description Length Principle}
MDL is one of many ways to measure complexity, including Kolmogorov complexity\citep{kolmogorov65, SOLOMONOFF19641}, VC dimension \citep{vapnik1971uniform} and Rademacher complexity \citep{bartlett2003rademacher}.
The Refined MDL coding scheme \citep{barron1998minimum} which defines stochastic complexity \citep{rissanen1986complexity} is based on the normalized maximum likliehood (NML) distribution \citep{universal1987shtarkov}, which has been modified for predictive settings (pNML) \citep{roos2008bayesian, fogel2019universal}.
Refined MDL bears similarity to model selection criteria such as Akaike Information Criterion (AIC) \citep{akaike1973information} and the Bayesian Information Criterion (BIC) \citep{10.1214/aos/1176344136}.
We point readers to \citet{grunwald2004tutorial} for further discussion of MDL.

MDL principles have been used to motivate methods to train autoencoders \citep{NIPS1993_9e3cfc48} and regularize weights \citep{10.1145/168304.168306}.
Applying MDL directly to overparameterized model classes is difficult, requiring subclass restrictions such as ridge estimators \citep{JMLR:v24:21-1133}.
Recent work in MDL methods for neural networks have mainly attempted to approximate pNML distributions by performing additional gradient steps \citep{bibas2020deep} or optimizing with an approximate posterior \citep{zhou2021amortized}.
Instead of explicitly calculating model weights, IF-COMP directly linearizes the model making optimization unnecessary and allowing post-hoc tuning of example weighting.

\paragraph{Quantifying Predictive Uncertainty} 
A wide variety of methods have been proposed to quantify predictive uncertainty in deep neural networks. 
Bayesian approaches include Laplace approximations \citep{mackay1992bayesian, ritter2018laplace}, variational inference \citep{NIPS2011_7eb3c8be}, stochastic gradient MCMC \citep{10.5555/3104482.3104568}, dropout \citep{gal2016dropout}, and weight averaging \citep{izmailov2019averaging, maddox2019swag}.
Non-Bayesian methods include model ensembling \citep{balaji2017ensembles, NIPS2016_8d8818c8}, Platt scaling \citep{PlattProbabilisticOutputs1999} with logit temperatures, \citep{pmlr-v70-guo17a} and more recently, MDL based approaches.

Bayesian and MDL methods have a deep connection.
For parametric hypothesis classes, MDL reduces to standard maximum likelihood estimation \citep{grunwald2004tutorial}.
Refined MDL model selection coincides with Bayes factor model selection \citep{kass1995bayes} based on a Jeffreys prior \citep{bernardo1994bayesian}, where codes correspond to MAP estimates. 
NML distributions over the full dataset can also be seen as a form of Bayesian marginal likelihood \citep{grunwald2004tutorial}, although pNML does not coincide with any standard Bayesian interpretations.

\paragraph{Influence Functions}
Influence functions \citep{75272a7e-1c8b-3ed5-9350-a7ee81abee59, bd831960-ac2b-396a-8c8f-de3944255f11} are a classical method from robust statistics that has found renewed interest in deep neural networks \citep{koh2017understanding, grosse2023studying}.
They have been applied for a wide range of tasks including detecting bias \citep{pmlr-v97-brunet19a}, data poisoning \citep{koh2021stronger}, auditing predictions \citep{pmlr-v89-schulam19a}, and fixing model mistakes \citep{NEURIPS2022_552260cf}.
The self-influence of a data point is related to its memorization \citep{feldman2021does}, which has been shown is necessary to cover the long tails of the data distribution and improve generalization \citep{feldman2020neural}. 
Although influence functions have been found to poorly match true retraining for neural networks \citep{basu2020influence}, they correlate well with an alternate proximal objective \eqref{eq:pbo} \citep{bae2022influence}.

\section{Conclusion}
In this paper we have proposed IF-COMP, an efficient method for estimating stochastic data complexity in deep neural networks
with temperature-scaled Boltzmann influence functions (BIFs).
The BIF softens the second order curvature and linearizes the model, allowing us to efficiently approximate hindsight-optimal outputs even for low probability labels.
Using these output estimates, IF-COMP can produce well calibrated output distributions as well as measure complexity on both labelled and unlabelled points.
On tasks covering uncertainty calibration, mislabel detection, and OOD detection, IF-COMP consistently matches or outperforms strong baselines, including Bayesian and optimization tracing based approaches.
Our results demonstrates the potential of MDL based approaches for improving uncertainty estimates in deep neural networks.

\section*{Acknowledgements}
Resources used in preparing this research were provided, in part, by the Province of Ontario, the Government of Canada through CIFAR, and companies sponsoring the Vector Institute \url{www.vectorinstitute.ai/#partners}.
We thank Vinith Suriyakumar, Neha Hulkund, and Kimia Hamidieh for helpful comments and discussion.

\bibliography{main}
\bibliographystyle{icml2024}

\newpage
\appendix
\onecolumn
\section{Additional Experiments}
\label{app:extra experiments}

\subsection{Hindsight-Optimal Retraining}
\label{app:unrestricted_oracle}
We consider ground truth retraining with the original hindsight objective \eqref{eq:inf_obj} on CIFAR-10 with a ResNet-18 model. Similar to our experiments in Section \ref{subsec:oracle_val}, we select 20 examples randomly from CIFAR-10, CIFAR-100, and MNIST test sets, then apply each of the possible 10 CIFAR labels for a total of 600 additional training examples.

For each example, we train the model \textit{from scratch} on the full training set with the added example, and evaluate the probability of the true label class at convergence. 
We find that the model is able to effectively memorize all additional examples provided to it, achieving close to 1 probability consistently across all classes.
Using this hindsight-optimal probability in the pNML formulation produces a constant uniform distribution that is useless for any tasks.

\subsection{Dataset Pruning}
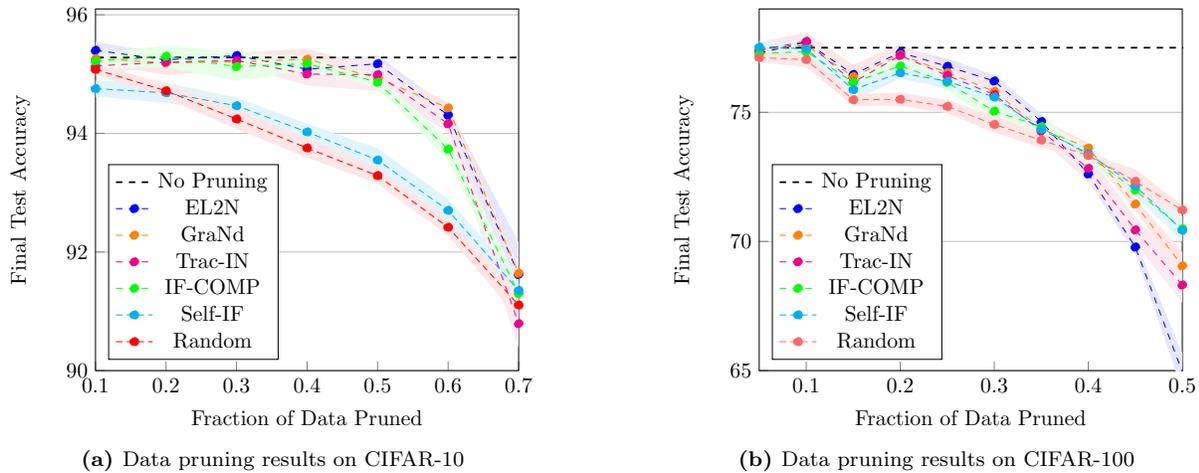
\begin{figure}
    \centering
    \subfloat[Data pruning results on CIFAR-10]{
    \resizebox{0.45\columnwidth}{!}{%
        \begin{tikzpicture}
\begin{axis}[
    xlabel=Fraction of Data Pruned,
    ylabel=Final Test Accuracy,
    xtick pos=left,
    ytick pos=left,
    ymajorgrids=true,
    xmin=0.1, xmax=0.7,
    ymin=90,
    height=7.5cm, width=8.5cm,
    legend pos=south west,
]

\addplot+[mark=none, black, thick, dashed, samples=1] {95.28};
\addlegendentry{No Pruning}

\addplot+[dashed, blue, mark options={blue}, mark=*] table [x=x, y=y, col sep=comma] {
x,  y
0.1, 95.4
0.2, 95.24
0.3, 95.312
0.4, 95.08200000000001
0.5, 95.174
0.6, 94.30799999999999
0.7, 91.62199999999999
};
\addlegendentry{EL2N}

\addplot [name path=el2n_lower,draw=none,forget plot] 
table[x=x,y=y, col sep=comma] {
x, y
0.1, 95.25647299905593
0.2, 95.14060181088168
0.3, 95.23417969416663
0.4, 94.96497008929339
0.5, 95.03291846329162
0.6, 94.17633375527494
0.7, 91.08046883746177
};

\addplot [name path=el2n_upper,draw=none,forget plot] 
table[x=x,y=y, col sep=comma] {
x, y
0.1, 95.54352700094408
0.2, 95.33939818911831
0.3, 95.38982030583337
0.4, 95.19902991070663
0.5, 95.31508153670839
0.6, 94.43966624472505
0.7, 92.1635311625382
};

\addplot [fill=blue!10,forget plot] fill between[of=el2n_upper and el2n_lower];

\addplot+[dashed, orange, mark options={orange}, mark=*] table [x=x, y=y, col sep=comma] {
x,  y
0.1, 95.25999999999999
0.2, 95.19
0.3, 95.20599999999999
0.4, 95.24999999999999
0.5, 94.935
0.6, 94.428
0.7, 91.642
};
\addlegendentry{GraNd}

\addplot [name path=grand_lower,draw=none,forget plot] 
table[x=x,y=y, col sep=comma] {
x, y
0.1, 95.17538321679476
0.2, 94.98800990123276
0.3, 95.05100322584
0.4, 95.06933456334981
0.5, 94.77106403689245
0.6, 94.32973505202769
0.7, 91.53666244734188
};

\addplot [name path=grand_upper,draw=none,forget plot] 
table[x=x,y=y, col sep=comma] {
x, y
0.1, 95.34461678320523
0.2, 95.39199009876724
0.3, 95.36099677415997
0.4, 95.43066543665016
0.5, 95.09893596310755
0.6, 94.52626494797231
0.7, 91.74733755265811
};

\addplot [fill=red!10,forget plot] fill between[of=grand_upper and grand_lower];

\addplot+[dashed, magenta, mark options={magenta}, mark=*] table [x=x, y=y, col sep=comma] {
x,  y
0.1, 95.13799999999999
0.2, 95.2
0.3, 95.23
0.4, 95.00000000000001
0.5, 94.99000000000001
0.6, 94.162
0.7, 90.79199999999999
};
\addlegendentry{Trac-IN}

\addplot [name path=tracin_lower,draw=none,forget plot] 
table[x=x,y=y, col sep=comma] {
x, y
0.1, 94.99248024189134
0.2, 95.13427329309938
0.3, 95.15075354897537
0.4, 94.80827102462071
0.5, 94.71745642550228
0.6, 94.02482857440414
0.7, 90.37894310319278
};

\addplot [name path=tracin_upper,draw=none,forget plot] 
table[x=x,y=y, col sep=comma] {
x, y
0.1, 95.28351975810864
0.2, 95.26572670690062
0.3, 95.30924645102463
0.4, 95.19172897537932
0.5, 95.26254357449774
0.6, 94.29917142559587
0.7, 91.20505689680719
};

\addplot [fill=magenta!10,forget plot] fill between[of=tracin_upper and tracin_lower];

\addplot+[green, dashed, mark options={green}, mark=*] table [x=x, y=y, col sep=comma] {
x,  y
0.1, 95.21600000000001
0.2, 95.3
0.3, 95.126
0.4, 95.17600000000002
0.5, 94.86200000000001
0.6, 93.734
0.7, 91.3
};
\addlegendentry{IF-COMP}

\addplot [name path=comp_lower,draw=none,forget plot] 
table[x=x,y=y, col sep=comma] {
x, y
0.1, 95.09359085001522
0.2, 95.11474342116945
0.3, 94.88611669503695
0.4, 95.06117839924475
0.5, 94.77495977941206
0.6, 93.54137601395465
0.7, 91.12100279331788
};

\addplot [name path=comp_upper,draw=none,forget plot] 
table[x=x,y=y, col sep=comma] {
x, y
0.1, 95.3384091499848
0.2, 95.48525657883054
0.3, 95.36588330496306
0.4, 95.29082160075528
0.5, 94.94904022058796
0.6, 93.92662398604534
0.7, 91.47899720668211
};

\addplot [fill=green!10,forget plot] fill between[of=comp_upper and comp_lower];

\addplot+[cyan, mark options={cyan}, mark=*] table [x=x, y=y, col sep=comma] {
x,  y
0.1, 94.75399999999999
0.2, 94.684
0.3, 94.466
0.4, 94.024
0.5, 93.55
0.6, 92.702
0.7, 91.35000000000001
};
\addlegendentry{Self-IF}

\addplot [name path=self_lower,draw=none,forget plot] 
table[x=x,y=y, col sep=comma] {
x, y
0.1, 94.61989556308608
0.2, 94.51118796338217
0.3, 94.33115193735169
0.4, 93.8853493599005
0.5, 93.35110304175276
0.6, 92.51359617838271
0.7, 91.15475656221015
};

\addplot [name path=self_upper,draw=none,forget plot] 
table[x=x,y=y, col sep=comma] {
x, y
0.1, 94.8881044369139
0.2, 94.85681203661782
0.3, 94.6008480626483
0.4, 94.1626506400995
0.5, 93.74889695824723
0.6, 92.89040382161728
0.7, 91.54524343778986
};

\addplot [fill=cyan!10,forget plot] fill between[of=self_upper and self_lower];

\addplot+[red, mark options={red}, mark=*] table [x=x, y=y, col sep=comma] {
x,  y
0.1, 95.078
0.2, 94.71799999999999
0.3, 94.242
0.4, 93.75399999999999
0.5, 93.28800000000001
0.6, 92.41600000000001
0.7, 91.106
};
\addlegendentry{Random}

\addplot [name path=random_lower,draw=none,forget plot] 
table[x=x,y=y, col sep=comma] {
x, y
0.1, 94.9895918555788
0.2, 94.59128772750833
0.3, 94.07378585077349
0.4, 93.59119950860024
0.5, 93.15303333744959
0.6, 92.27520227274563
0.7, 90.8760782741888
};

\addplot [name path=random_upper,draw=none,forget plot] 
table[x=x,y=y, col sep=comma] {
x, y
0.1, 95.16640814442121
0.2, 94.84471227249165
0.3, 94.41021414922652
0.4, 93.91680049139974
0.5, 93.42296666255044
0.6, 92.55679772725439
0.7, 91.33592172581119
};

\addplot [fill=red!10,forget plot] fill between[of=random_upper and random_lower];

\end{axis}
\end{tikzpicture}
    }}
    \hfill
    \subfloat[Data pruning results on CIFAR-100]{
    \resizebox{0.45\columnwidth}{!}{%
        \begin{tikzpicture}
\begin{axis}[
    xlabel=Fraction of Data Pruned,
    ylabel=Final Test Accuracy,
    xtick pos=left,
    ytick pos=left,
    ymajorgrids=true,
    xmin=0.05, xmax=0.5,
    ymin=65, ymax=79,
    height=7.5cm, width=8.5cm,
    legend pos=south west,
]

\addplot+[mark=none, black, thick, dashed, samples=1] {77.50};
\addlegendentry{No Pruning}

\addplot+[dashed, blue, mark options={blue}, mark=*] table [x=x, y=y, col sep=comma] {
x,  y
0.05, 77.284
0.1, 77.718
0.15, 76.466
0.2, 77.314
0.25, 76.77799999999999
0.3, 76.21000000000001
0.35, 74.64000000000001
0.4, 72.614
0.45, 69.7825
0.5, 64.952
};
\addlegendentry{EL2N}

\addplot [name path=el2n_lower,draw=none,forget plot] 
table[x=x,y=y, col sep=comma] {
x, y
0.05, 77.03737477825656
0.1, 77.35543138580401
0.15, 76.11961293326684
0.2, 77.10115263684978
0.25, 76.47183664490993
0.3, 75.97917539125996
0.35, 74.3290659233857
0.4, 72.3217261558059
0.45, 69.51722891224259
0.5, 64.22982550585056
};

\addplot [name path=el2n_upper,draw=none,forget plot] 
table[x=x,y=y, col sep=comma] {
x, y
0.05, 77.53062522174345
0.1, 78.080568614196
0.15, 76.81238706673315
0.2, 77.52684736315021
0.25, 77.08416335509006
0.3, 76.44082460874006
0.35, 74.95093407661433
0.4, 72.9062738441941
0.45, 70.04777108775741
0.5, 65.67417449414944
};

\addplot [fill=blue!10,forget plot] fill between[of=el2n_upper and el2n_lower];

\addplot+[dashed, orange, mark options={orange}, mark=*] table [x=x, y=y, col sep=comma] {
x,  y
0.05, 77.41799999999999
0.1, 77.594
0.15, 76.38000000000001
0.2, 77.224
0.25, 76.538
0.3, 75.80799999999999
0.35, 74.35799999999999
0.4, 73.618
0.45, 71.446
0.5, 69.054
};
\addlegendentry{GraNd}

\addplot [name path=grand_lower,draw=none,forget plot] 
table[x=x,y=y, col sep=comma] {
x, y
0.05, 77.12855915975798
0.1, 77.09485673399313
0.15, 75.92762847127611
0.2, 76.79455384505157
0.25, 76.21985537879762
0.3, 75.45188765817511
0.35, 73.99731176897491
0.4, 73.41040423896428
0.45, 71.2208245128794
0.5, 68.51926268130978
};

\addplot [name path=grand_upper,draw=none,forget plot] 
table[x=x,y=y, col sep=comma] {
x, y
0.05, 77.707440840242
0.1, 78.09314326600686
0.15, 76.83237152872391
0.2, 77.65344615494844
0.25, 76.85614462120238
0.3, 76.16411234182488
0.35, 74.71868823102507
0.4, 73.82559576103571
0.45, 71.67117548712059
0.5, 69.58873731869022
};

\addplot [fill=red!10,forget plot] fill between[of=grand_upper and grand_lower];

\addplot+[dashed, magenta, mark options={magenta}, mark=*] table [x=x, y=y, col sep=comma] {
x,  y
0.05, 77.402
0.1, 77.74600000000001
0.15, 76.162
0.2, 77.196
0.25, 76.422
0.3, 75.684
0.35, 74.272
0.4, 72.83200000000001
0.45, 70.44800000000001
0.5, 68.314
};
\addlegendentry{Trac-IN}

\addplot [name path=tracin_lower,draw=none,forget plot] 
table[x=x,y=y, col sep=comma] {
x, y
0.05, 77.1922000953289
0.1, 77.53646718634067
0.15, 75.85832912553226
0.2, 76.95255801512475
0.25, 76.12310871541645
0.3, 75.42410771462008
0.35, 73.92286392337658
0.4, 72.3305062313448
0.45, 69.94340511298667
0.5, 67.63201466291422
};

\addplot [name path=tracin_upper,draw=none,forget plot] 
table[x=x,y=y, col sep=comma] {
x, y
0.05, 77.6117999046711
0.1, 77.95553281365935
0.15, 76.46567087446775
0.2, 77.43944198487524
0.25, 76.72089128458354
0.3, 75.94389228537992
0.35, 74.62113607662343
0.4, 73.33349376865522
0.45, 70.95259488701335
0.5, 68.99598533708577
};

\addplot [fill=magenta!10,forget plot] fill between[of=tracin_upper and tracin_lower];

\addplot+[green, dashed, mark options={green}, mark=*] table [x=x, y=y, col sep=comma] {
x,  y
0.05, 77.27000000000001
0.1, 77.35
0.15, 76.158
0.2, 76.79400000000001
0.25, 76.15800000000002
0.3, 75.04400000000001
0.35, 74.446
0.4, 73.334
0.45, 71.98799999999999
0.5, 70.48799999999999
};
\addlegendentry{IF-COMP}

\addplot [name path=comp_lower,draw=none,forget plot] 
table[x=x,y=y, col sep=comma] {
x, y
0.05, 76.91155893092449
0.1, 76.98105555973832
0.15, 75.90129783795223
0.2, 76.48232067761882
0.25, 75.86058446577223
0.3, 74.5925755877226
0.35, 74.15751429844791
0.4, 73.04468356424151
0.45, 71.76660442642184
0.5, 69.99264558142679
};

\addplot [name path=comp_upper,draw=none,forget plot] 
table[x=x,y=y, col sep=comma] {
x, y
0.05, 77.62844106907554
0.1, 77.71894444026167
0.15, 76.41470216204777
0.2, 77.1056793223812
0.25, 76.4554155342278
0.3, 75.49542441227742
0.35, 74.73448570155209
0.4, 73.6233164357585
0.45, 72.20939557357813
0.5, 70.98335441857319
};

\addplot [fill=green!10,forget plot] fill between[of=comp_upper and comp_lower];

\addplot+[cyan, mark options={cyan}, mark=*] table [x=x, y=y, col sep=comma] {
x,  y
0.05, 77.52000000000001
0.1, 77.462
0.15, 75.87800000000001
0.2, 76.526
0.25, 76.2
0.3, 75.59
0.35, 74.34
0.4, 73.4
0.45, 72.148
0.5, 70.436
};
\addlegendentry{Self-IF}

\addplot [name path=self_lower,draw=none,forget plot] 
table[x=x,y=y, col sep=comma] {
x, y
0.05, 77.28016672457727
0.1, 77.22770104567029
0.15, 75.44296436927536
0.2, 76.30813765814166
0.25, 75.92606570130778
0.3, 75.33836534419918
0.35, 74.18469385073347
0.4, 73.18228458942923
0.45, 71.76327152431877
0.5, 70.03397512514772
};

\addplot [name path=self_upper,draw=none,forget plot] 
table[x=x,y=y, col sep=comma] {
x, y
0.05, 77.75983327542275
0.1, 77.69629895432972
0.15, 76.31303563072467
0.2, 76.74386234185833
0.25, 76.47393429869223
0.3, 75.84163465580083
0.35, 74.49530614926654
0.4, 73.61771541057078
0.45, 72.53272847568122
0.5, 70.8380248748523
};

\addplot [fill=cyan!10,forget plot] fill between[of=self_upper and self_lower];

\addplot+[red!60, mark options={red!60}, mark=*] table [x=x, y=y, col sep=comma] {
x,  y
0.05, 77.114
0.1, 77.03999999999999
0.15, 75.484
0.2, 75.502
0.25, 75.22600000000001
0.3, 74.53
0.35, 73.922
0.4, 73.328
0.45, 72.326
0.5, 71.21799999999999
};
\addlegendentry{Random}

\addplot [name path=random_lower,draw=none,forget plot] 
table[x=x,y=y, col sep=comma] {
x, y
0.05, 76.94304971482914
0.1, 76.74086792214808
0.15, 75.24303527229073
0.2, 75.24961933513043
0.25, 74.96580007686397
0.3, 74.21682273390299
0.35, 73.62304180894313
0.4, 73.25605557700558
0.45, 71.85548538811211
0.5, 71.04440276499896
};

\addplot [name path=random_upper,draw=none,forget plot] 
table[x=x,y=y, col sep=comma] {
x, y
0.05, 77.28495028517086
0.1, 77.3391320778519
0.15, 75.72496472770926
0.2, 75.75438066486956
0.25, 75.48619992313606
0.3, 74.84317726609702
0.35, 74.22095819105687
0.4, 73.39994442299442
0.45, 72.79651461188787
0.5, 71.39159723500102
};

\addplot [fill=red!10,forget plot] fill between[of=random_upper and random_lower];

\end{axis}
\end{tikzpicture}
    }}
\caption{Data pruning results. Shaded regions correspond to standard deviations over 5 seeds.IF-COMP performs similarly to other methods that require access to additional checkpoints, including Trac-IN, GraNd, and EL2N. At the highest pruning levels for CIFAR-100, IF-COMP and Self-IF outperform baselines that perform worse than random.}
\label{fig:app_pruning}
\end{figure}

We consider the task of dataset pruning, with the goal of removing training examples that do not provide essential training information to the model.
Specifically, given a model trained on a dataset, we aim to rank training examples such that removing the least important examples degrades final test accuracy as little as possible.
We consider ResNet-18 models trained on both CIFAR-10 and CIFAR-100 datasets, with varying levels of data pruning.
We use IF-COMP to rank points, removing ones with the lowest complexity first and keeping those with higher complexity.

For baselines, we consider Trac-IN, GraNd, EL2N, and Self-IF, which follow our experimental setup for mislabel detection.
We also consider a no pruning and random pruning baseline.
Once points are ranked, we train 5 models with different seeds on each pruned dataset.
Results are shown in Figure \ref{fig:app_pruning}.
We find that IF-COMP performs similarly to baseline methods across both datasets, even those that utilize additional training checkpoints.
On CIFAR-100 at the highest pruning levels, IF-COMP and Self-IF outperform baselines, although all methods perform worse than random.

\subsection{$\alpha$ and $\beta$ Ablations}
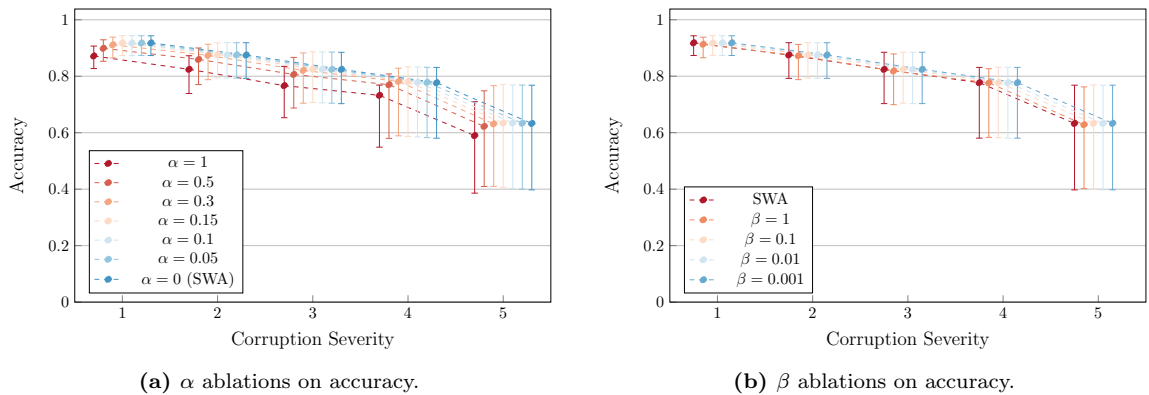
\begin{figure*}
    \centering
    \subfloat[$\alpha$ ablations on accuracy.]{\centering\resizebox{0.46\columnwidth}{!}{%
        \pgfplotsset{
    cycle list/RdBu-8,
}

\begin{tikzpicture}
\begin{axis}[
    label style={align=center,font=\large},
    xlabel=Corruption Severity,
    ylabel=Accuracy,
    xtick pos=left,
    xtick={1,2,3,4,5},
    ytick pos=left,
    ymajorgrids=true,
    xmin=0.5,xmax=5.5,
    ymin=0,
    height=8cm, width=12cm,
    legend pos=south west,
    yticklabel style={
        /pgf/number format/fixed,
        /pgf/number format/precision=5
    },
    legend style={column sep=10pt,},
    scaled y ticks=false,
]

\addplot+[mark=*, dashed, error bars/.cd, error bar style={solid}, y dir=both, y explicit] table [x=x, y=y,y error plus=plus, y error minus=minus, col sep=comma] {
x,  y, minus, plus
0.7, 0.8718, 0.044600000000000084, 0.03510000000000002
1.7, 0.8246, 0.08565, 0.04800000000000004
2.7, 0.7667, 0.11360000000000003, 0.06829999999999992
3.7, 0.7324, 0.18364999999999998, 0.03639999999999999
4.7, 0.5897, 0.20375, 0.12019999999999997
};
\addlegendentry{$\alpha = 1$}

\addplot+[mark=*, dashed, error bars/.cd, y dir=both, y explicit, error bar style={solid}] table [x=x, y=y,y error plus=plus, y error minus=minus, col sep=comma] {
x,  y, minus, plus
0.8, 0.899, 0.04590000000000005, 0.030100000000000016
1.8, 0.8596, 0.08900000000000008, 0.04049999999999998
2.8, 0.8065, 0.11904999999999999, 0.06030000000000002
3.8, 0.77, 0.19000000000000006, 0.038250000000000006
4.8, 0.6223, 0.21275, 0.1261500000000001
};
\addlegendentry{$\alpha = 0.5$}

\addplot+[mark=*, dashed, error bars/.cd, error bar style={solid}, y dir=both, y explicit] table [x=x, y=y,y error plus=plus, y error minus=minus, col sep=comma] {
x,  y, minus, plus
0.9, 0.9115, 0.04634999999999989, 0.027150000000000007
1.9, 0.8739, 0.08594999999999997, 0.039449999999999985
2.9, 0.8207, 0.11609999999999998, 0.06194999999999995
3.9, 0.7817, 0.19274999999999998, 0.046800000000000064
4.9, 0.6311, 0.22109999999999996, 0.13549999999999995
};
\addlegendentry{$\alpha = 0.3$}

\addplot+[mark=*, dashed, error bars/.cd, error bar style={solid}, y dir=both, y explicit] table [x=x, y=y,y error plus=plus, y error minus=minus, col sep=comma] {
x,  y, minus, plus
1.0, 0.918, 0.04435, 0.02574999999999994
2.0, 0.8773, 0.0825499999999999, 0.041100000000000025
3.0, 0.8259, 0.11885000000000001, 0.06074999999999997
4.0, 0.7805, 0.19389999999999996, 0.05315000000000003
5.0, 0.6342, 0.22775, 0.13685000000000003
};
\addlegendentry{$\alpha = 0.15$}

\addplot+[mark=*, dashed, error bars/.cd, error bar style={solid}, y dir=both, y explicit] table [x=x, y=y,y error plus=plus, y error minus=minus, col sep=comma] {
x,  y, minus, plus
1.1, 0.9177, 0.043849999999999945, 0.026000000000000023
2.1, 0.876, 0.08195000000000008, 0.04304999999999992
3.1, 0.8257, 0.11985000000000001, 0.060350000000000015
4.1, 0.779, 0.19395000000000007, 0.053649999999999975
5.1, 0.6345, 0.23169999999999996, 0.13500000000000012
};
\addlegendentry{$\alpha = 0.1$}

\addplot+[mark=*, dashed, error bars/.cd, error bar style={solid}, y dir=both, y explicit] table [x=x, y=y,y error plus=plus, y error minus=minus, col sep=comma] {
x,  y, minus, plus
1.2, 0.9177, 0.043949999999999934, 0.025800000000000045
2.2, 0.8757, 0.08219999999999994, 0.043200000000000016
3.2, 0.8245, 0.11995, 0.06125000000000003
4.2, 0.7782, 0.19574999999999998, 0.05349999999999999
5.2, 0.6336, 0.23380000000000006, 0.1350999999999999
};
\addlegendentry{$\alpha = 0.05$}

\addplot+[mark=*, dashed, error bars/.cd, error bar style={solid}, y dir=both, y explicit] table [x=x, y=y,y error plus=plus, y error minus=minus, col sep=comma] {
x,  y, minus, plus
1.3, 0.9181, 0.04480000000000006, 0.025149999999999895
2.3, 0.8752, 0.08279999999999998, 0.04354999999999998
3.3, 0.8243, 0.12135000000000007, 0.0605
4.3, 0.7771, 0.1965, 0.05410000000000004
5.3, 0.6328, 0.23535000000000006, 0.13539999999999996
};
\addlegendentry{$\alpha = 0$ (SWA)}



\end{axis}
\end{tikzpicture}
    }\label{subfig:alpha_acc}}
    \hfil
    \subfloat[$\beta$ ablations on accuracy.]{\centering\resizebox{0.46\columnwidth}{!}{%
        \pgfplotsset{
    cycle list/RdBu-6,
}

\begin{tikzpicture}
\begin{axis}[
    label style={align=center,font=\large},
    xlabel=Corruption Severity,
    ylabel=Accuracy,
    xtick pos=left,
    xtick={1,2,3,4,5},
    ytick pos=left,
    ymajorgrids=true,
    xmin=0.5,xmax=5.5,
    ymin=0,
    height=8cm, width=12cm,
    legend pos=south west,
    yticklabel style={
        /pgf/number format/fixed,
        /pgf/number format/precision=5
    },
    legend style={column sep=10pt,},
    scaled y ticks=false,
]

\addplot+[mark=*, dashed, error bars/.cd, y dir=both, y explicit, error bar style={solid}] table [x=x, y=y,y error plus=plus, y error minus=minus, col sep=comma] {
x,  y, minus, plus
0.75, 0.9181, 0.04480000000000006, 0.025149999999999895
1.75, 0.8752, 0.08279999999999998, 0.04354999999999998
2.75, 0.8243, 0.12135000000000007, 0.0605
3.75, 0.7771, 0.1965, 0.05410000000000004
4.75, 0.6328, 0.23535000000000006, 0.13539999999999996
};
\addlegendentry{SWA}

\addplot+[mark=*, dashed, error bars/.cd, y dir=both, y explicit, error bar style={solid}] table [x=x, y=y,y error plus=plus, y error minus=minus, col sep=comma] {
x,  y, minus, plus
0.85, 0.9132, 0.04794999999999994, 0.025150000000000006
1.85, 0.8725, 0.08465000000000011, 0.04014999999999991
2.85, 0.8185, 0.11935000000000007, 0.060349999999999904
3.85, 0.7768, 0.19340000000000013, 0.04964999999999997
4.85, 0.6283, 0.22585, 0.13395
};
\addlegendentry{$\beta = 1$}

\addplot+[mark=*, dashed, error bars/.cd, error bar style={solid}, y dir=both, y explicit] table [x=x, y=y,y error plus=plus, y error minus=minus, col sep=comma] {
x,  y, minus, plus
0.95, 0.9178, 0.0442499999999999, 0.02575000000000005
1.95, 0.8756, 0.08189999999999997, 0.04314999999999991
2.95, 0.8247, 0.12034999999999996, 0.06075000000000008
3.95, 0.778, 0.19579999999999997, 0.05345
4.95, 0.6331, 0.23370000000000002, 0.13559999999999994
};
\addlegendentry{$\beta = 0.1$}

\addplot+[mark=*, dashed, error bars/.cd, error bar style={solid}, y dir=both, y explicit] table [x=x, y=y,y error plus=plus, y error minus=minus, col sep=comma] {
x,  y, minus, plus
1.05, 0.9181, 0.044849999999999945, 0.025249999999999995
2.05, 0.8753, 0.08274999999999988, 0.04354999999999998
3.05, 0.8244, 0.12139999999999995, 0.0605
4.05, 0.7773, 0.19665, 0.05415000000000003
5.05, 0.6329, 0.23535000000000006, 0.13535000000000008
};
\addlegendentry{$\beta = 0.01$}

\addplot+[mark=*, dashed, error bars/.cd, error bar style={solid}, y dir=both, y explicit] table [x=x, y=y,y error plus=plus, y error minus=minus, col sep=comma] {
x,  y, minus, plus
1.15, 0.9181, 0.04480000000000006, 0.025149999999999895
2.15, 0.8752, 0.08279999999999998, 0.04354999999999998
3.15, 0.8244, 0.12139999999999995, 0.06040000000000001
4.15, 0.7771, 0.1965, 0.05415000000000003
5.15, 0.6329, 0.23535, 0.13529999999999998
};
\addlegendentry{$\beta = 0.001$}



\end{axis}
\end{tikzpicture}
    }\label{subfig:beta_acc}}
    \caption{Ablations on the effects of the $\alpha$ and $\beta$ hyperparameters on model accuracy. IF-COMP exhibits minimal differences from SWA baseline accuracy at all severity levels, except for particularly large values of $\alpha$.}
    \label{fig:abl_acc}
\end{figure*}

\begin{figure*}
    \centering
    \subfloat[$\alpha$ ablations on ECE]{\centering\resizebox{0.46\columnwidth}{!}{%
        \pgfplotsset{
    cycle list/RdBu-8,
}

\begin{tikzpicture}
\begin{axis}[
    label style={align=center,font=\large},
    xlabel=Corruption Severity,
    ylabel=ECE,
    xtick pos=left,
    xtick={1,2,3,4,5},
    ytick pos=left,
    ymajorgrids=true,
    xmin=0.5,xmax=5.5,
    ymin=0,
    height=8cm, width=12cm,
    legend pos=north west,
    yticklabel style={
        /pgf/number format/fixed,
        /pgf/number format/precision=5
    },
    legend style={column sep=10pt,},
    scaled y ticks=false,
]

\addplot+[mark=*, only marks, dashed, error bars/.cd, error bar style={solid}, y dir=both, y explicit] table [x=x, y=y,y error plus=plus, y error minus=minus, col sep=comma] {
x,  y, minus, plus
0.7, 0.04332983801364897, 0.004354905021190618, 0.0015722521245479723
1.7, 0.04604167190790177, 0.004446424305438998, 0.00243523732423781
2.7, 0.05105625145435333, 0.00590130299329758, 0.00710278691649438
3.7, 0.05448448004722595, 0.01257820814251899, 0.026746971994638435
4.7, 0.05442881551980972, 0.008682141613960269, 0.061793802016973494
};
\addlegendentry{$\alpha = 1$}

\addplot+[mark=*, only marks, dashed, error bars/.cd, y dir=both, y explicit, error bar style={solid}] table [x=x, y=y,y error plus=plus, y error minus=minus, col sep=comma] {
x,  y, minus, plus
0.8, 0.03451231620311736, 0.004053289735317218, 0.0031767191052436833
1.8, 0.03652313270568848, 0.0047001030802726815, 0.007264344334602345
2.8, 0.03989804570674896, 0.003940133374929426, 0.01934717680811883
3.8, 0.04930740119218827, 0.015525467622280131, 0.02388701173067092
4.8, 0.041724608600139615, 0.0066520569801330615, 0.08780474945306778
};
\addlegendentry{$\alpha = 0.5$}

\addplot+[mark=*, only marks, dashed, error bars/.cd, error bar style={solid}, y dir=both, y explicit] table [x=x, y=y,y error plus=plus, y error minus=minus, col sep=comma] {
x,  y, minus, plus
0.9, 0.02923272823095321, 0.0016997323274612376, 0.0027365757584571805
1.9, 0.03292842284440994, 0.006742728173732754, 0.005138693362474446
2.9, 0.03254452562332154, 0.004606034159660346, 0.014501329094171511
3.9, 0.03753963534832, 0.006236819297075265, 0.03991403827667238
4.9, 0.03814069777727127, 0.009706323134899136, 0.11264057971239089
};
\addlegendentry{$\alpha = 0.3$}

\addplot+[mark=*, only marks, dashed, error bars/.cd, error bar style={solid}, y dir=both, y explicit] table [x=x, y=y,y error plus=plus, y error minus=minus, col sep=comma] {
x,  y, minus, plus
1.0, 0.01888017148971557, 0.0011697871983051313, 0.003129631769657134
2.0, 0.018065453612804414, 0.0019906804025173197, 0.0076568695843219695
3.0, 0.018468522918224334, 0.0022592093944549564, 0.017104474216699608
4.0, 0.019424964034557346, 0.003133470559120173, 0.09020849146246912
5.0, 0.054936809873580926, 0.030958996695280065, 0.13733831329941745
};
\addlegendentry{$\alpha = 0.15$}

\addplot+[mark=*, only marks, dashed, error bars/.cd, error bar style={solid}, y dir=both, y explicit] table [x=x, y=y,y error plus=plus, y error minus=minus, col sep=comma] {
x,  y, minus, plus
1.1, 0.012514838469028466, 0.0009897493600845245, 0.00046651906967163856
2.1, 0.015064835834503165, 0.004358744454383848, 0.009012045955657974
3.1, 0.015028689694404613, 0.0035053539037704622, 0.041152535623311996
4.1, 0.022367194318771353, 0.008891104990243905, 0.1123020413219929
5.1, 0.08032961032390593, 0.04373816968202589, 0.1347943173527718
};
\addlegendentry{$\alpha = 0.1$}

\addplot+[mark=*, only marks, dashed, error bars/.cd, error bar style={solid}, y dir=both, y explicit] table [x=x, y=y,y error plus=plus, y error minus=minus, col sep=comma] {
x,  y, minus, plus
1.2, 0.007879364657402032, 0.0006560122251510567, 0.0101013726770878
2.2, 0.013018887698650358, 0.004083129590749737, 0.032880278384685525
3.2, 0.024613580048084255, 0.011001312983036039, 0.05776447708010676
4.2, 0.04281046545505524, 0.02042620766162873, 0.12243209481835365
5.2, 0.11151196395158766, 0.05552867861390112, 0.132231876540184
};
\addlegendentry{$\alpha = 0.05$}

\addplot+[mark=*, only marks, dashed, error bars/.cd, error bar style={solid}, y dir=both, y explicit] table [x=x, y=y,y error plus=plus, y error minus=minus, col sep=comma] {
x,  y, minus, plus
1.3, 0.01738921004533767, 0.004345595419406887, 0.019786821234226235
2.3, 0.030144979274272926, 0.010724680298566824, 0.04657674098014833
3.3, 0.04982058678865434, 0.024298554646968853, 0.06791076707243918
4.3, 0.06924911025762558, 0.02281877567768096, 0.1367949703633785
5.3, 0.15109313688278198, 0.06630255128145217, 0.13008213880062108
};
\addlegendentry{$\alpha = 0$ (SWA + Temp)}



\end{axis}
\end{tikzpicture}
    }\label{subfig:alpha_ece}}
    \hfil
    \subfloat[$\beta$ ablations on ECE]{\centering\resizebox{0.46\columnwidth}{!}{%
        \pgfplotsset{
    cycle list/RdBu-6,
}

\begin{tikzpicture}
\begin{axis}[
    label style={align=center,font=\large},
    xlabel=Corruption Severity,
    ylabel=ECE,
    xtick pos=left,
    xtick={1,2,3,4,5},
    ytick pos=left,
    ymajorgrids=true,
    xmin=0.5,xmax=5.5,
    ymin=0,
    height=8cm, width=12cm,
    legend pos=north west,
    yticklabel style={
        /pgf/number format/fixed,
        /pgf/number format/precision=5
    },
    legend style={column sep=10pt,},
    scaled y ticks=false,
]

\addplot+[mark=*, dashed, error bars/.cd, y dir=both, y explicit, error bar style={solid}] table [x=x, y=y,y error plus=plus, y error minus=minus, col sep=comma] {
x,  y, minus, plus
0.75, 0.053920411491394045, 0.016420160031318666, 0.03509076099395755
1.75, 0.08237901668548583, 0.02754628231525419, 0.06828492825031285
2.75, 0.11574640132188797, 0.040584225535392746, 0.09564685294628149
3.75, 0.14810351560115814, 0.03308053835630416, 0.16817066751122478
4.75, 0.2790144944310189, 0.11639362618923194, 0.12383122318983064
};
\addlegendentry{SWA ($\beta = 1.25$)}

\addplot+[mark=*, dashed, error bars/.cd, y dir=both, y explicit, error bar style={solid}] table [x=x, y=y,y error plus=plus, y error minus=minus, col sep=comma] {
x,  y, minus, plus
0.85, 0.026739547836780553, 0.006362178164720535, 0.022564531093835837
1.85, 0.039257750940322876, 0.010583607232570649, 0.04880668219923973
2.85, 0.05695278116464616, 0.01905405883193017, 0.06430269818305967
3.85, 0.07927080202102663, 0.022150103455781944, 0.12443615489602089
4.85, 0.1529240477323532, 0.05621664552688599, 0.13017705938220028
};
\addlegendentry{IF-COMP ($\beta = 1.25$)}

\addplot+[mark=*, dashed, error bars/.cd, error bar style={solid}, y dir=both, y explicit] table [x=x, y=y,y error plus=plus, y error minus=minus, col sep=comma] {
x,  y, minus, plus
0.95, 0.04542517008781434, 0.013815651631355286, 0.0319016263961792
1.95, 0.06936484615802767, 0.022687787020206462, 0.06327539776563644
2.95, 0.09830671854019164, 0.035056031823158265, 0.0889741382956505
3.95, 0.12879424670934678, 0.03029258176088334, 0.1616161131083965
4.95, 0.2485026296138763, 0.10630479148030281, 0.11640727835297576
};
\addlegendentry{SWA ($\beta = 1$)}

\addplot+[mark=*, dashed, error bars/.cd, error bar style={solid}, y dir=both, y explicit] table [x=x, y=y,y error plus=plus, y error minus=minus, col sep=comma] {
x,  y, minus, plus
1.05, 0.016425154387950905, 0.005776135367155079, 0.015263532978296274
2.05, 0.026024000763893135, 0.009591396981477741, 0.03376236483454703
3.05, 0.03558119325637818, 0.012087151318788535, 0.05450271927118301
4.05, 0.05448615039587021, 0.019372231680154797, 0.11101596285700796
5.05, 0.11564786868095397, 0.044687354958057396, 0.1333402727842331
};
\addlegendentry{IF-COMP ($\beta = 1$)}

\addplot+[mark=*, dashed, error bars/.cd, error bar style={solid}, y dir=both, y explicit] table [x=x, y=y,y error plus=plus, y error minus=minus, col sep=comma] {
x,  y, minus, plus
1.15, 0.017389207065105435, 0.004345574557781218, 0.019786826449632648
2.15, 0.030144979274272926, 0.010723449462652211, 0.04657674098014833
3.15, 0.049820595729351054, 0.0242973282814026, 0.06791075962185861
4.15, 0.0692491191983223, 0.022818787598609926, 0.13679496589303017
5.15, 0.15109313688278198, 0.06630255202651024, 0.13008214625120157
};
\addlegendentry{SWA ($\beta = 0.66$)}

\addplot+[mark=*, dashed, error bars/.cd, error bar style={solid}, y dir=both, y explicit] table [x=x, y=y,y error plus=plus, y error minus=minus, col sep=comma] {
x,  y, minus, plus
1.25, 0.01908738873004913, 0.0009846703767776473, 0.003344981658458715
2.25, 0.018639648795127865, 0.0021044900953769673, 0.007724683380126954
3.25, 0.019135190057754506, 0.0026124103903770388, 0.015592345535755184
4.25, 0.01884255959987641, 0.0023831518471241, 0.08941039566397666
5.25, 0.05339602445363998, 0.029645656210184085, 0.1376670102596283
};
\addlegendentry{IF-COMP ($\beta = 0.66$)}



\end{axis}
\end{tikzpicture}
    }\label{subfig:beta_ece}}
    \caption{Ablations on the effects of the $\alpha$ and $\beta$ hyperparameters on ECE. Different values of $\alpha$ are optimal for different severity levels. Selecting an $\alpha$ value allows practitioners to prioritize calibration on different images. Applying IF-COMP on top of SWA models improves ECE across all severity levels even for varying values of $\beta$ ($\alpha$ is not tuned and held constant at 0.15).}
    \label{fig:abl_ece}
\end{figure*}
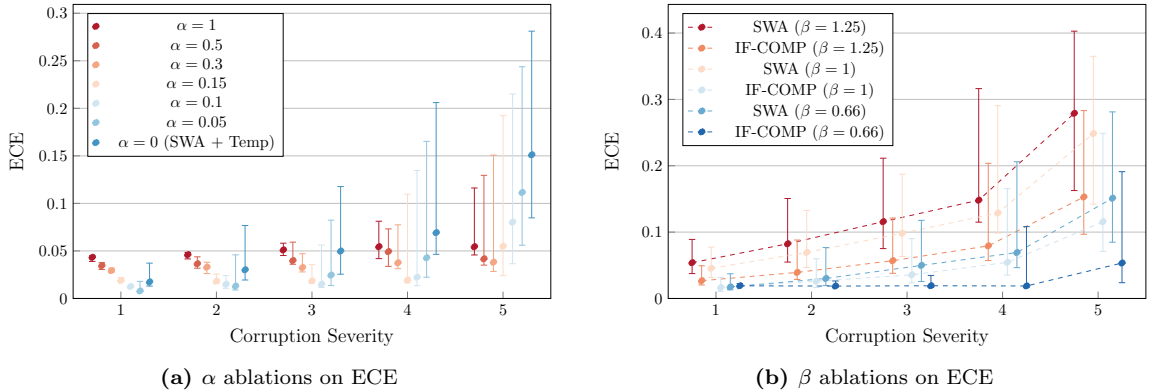

In this section we provide additional ablations on the $\alpha$ weighting parameter and $\beta$ inverse temperature parameter for the uncertainty calibration task.
We measure their effect on accuracy in Figure \ref{fig:abl_acc} and on ECE in Figure \ref{fig:abl_ece}.
We compare all methods against the SWA baseline.
When not specified, we set default values of $\beta = 0.66$ and $\alpha = 0.15$.

We find that accuracy is minimally affected by $\alpha$, and only begins to slightly degrade as $\alpha$ becomes particularly large ($> 0.5$). Since a maximum value of only $\alpha=0.3$ is required to achieve optimal ECE on severity level 5, even prioritizing calibration on the highest severity levels does not degrade accuracy.
Similarly, $\beta$ values have little effect on accuracy, even for extremely small values. This indicates that the BIF for each label provides as much information as the model output.

We find that the ECE for each corruption severity level changes with $\alpha$, where smaller values of $\alpha$ are better for cleaner images, and larger values of $\alpha$ are better for noisier images. 
Tuning $\alpha$ allows practitioners to select what kinds of test images to prioritize strong calibration on.
Compared to the raw outputs produced by temperature scaling SWA model outputs, IF-COMP consistently improves ECE across all severity levels for varying values of $\beta$. We do not tune a separate $\alpha$ for each $\beta$, and use the same value of $0.15$.

\section{Experimental Details}
\label{app:experimental_details}
\subsection{Computing Environment}
All experiments were implemented in PyTorch and were run on single RTX6000 or A40 GPUs.

\subsection{EKFAC}
We implement a version of the Eigenvalue Kronecker-Factored Approximate Curvature (EKFAC) \citep{george2021fast} to efficiently approximate the inverse Fisher information matrix.
For efficiency, in convolutional layers we assume spatially uncorrelated activations (SUA), except for MNIST LeNet models, for which removing the SUA assumption does not introduce significant computational overhead.
BatchNorm layers are implemented with a diagonal approximation which do not require eigendecomposition.
We take a single pass through the full training set and sample a single label to calculate the activation and output gradient covariances, then perform the relevant eigendecompositions.
We then take another full pass through the training set to estimate the second moments in the eigenbasis with a single label sample.
Both passes perform no data augmentation and use the original training images.
Calculating the BIF is performed by taking a temperature-scaled loss gradient, projecting into the eigenbasis, then taking the norm scaled by the inverse of the second moments.
We apply a consistent damping value of $\delta = 1\mathrm{e}{-30}$ for LeNet models and $\delta = 1\mathrm{e}{-12}$ for ResNet-18 models.

To further improve efficiency, we use the PyTorch \texttt{vmap} operation to vectorize computations both across label spaces and across examples. We use an example chunk size of 8 for MNIST and CIFAR-10 experiments, and an example chunk size of 4 for CIFAR-100 experiments, with full chunking across the label space. 

\subsection{Uncertainty Calibration}
All models are ResNet-18 models. 
CIFAR-10 ensemble models were trained with following standard training procedures using SGD with momentum of 0.9, weight decay of 0.0005, and a learning rate of 0.1 that decays by a factor of 5 at epochs 60, 120, and 160.
To train SWA models we follow the setup from \citep{izmailov2019averaging} and follow this setup for 160 epochs, then set a constant 0.01 learning rate and average checkpoints for another 140 epochs for a total of 300 epochs.
For SWA and SWAG-D models, batch statistics were then updated on the full training set.
For dropout models, we place dropout layers after the ReLU nonlinearities in each block before the residual connection, and test rates of 0.1, 0.2, 0.3, and 0.4. Empirically we find a rate of 0.3 to perform the best without significant accuracy degradation.
We follow the setup described in \citet{zhou2021amortized} for training ACNML models, including the recommended step size and weighting term.
All Bayesian methods use 30 samples from the posterior which we average.
IF-COMP uses a temperature of 1.5 and an $\alpha$ weight of $0.15$, which we find by tuning on a held out validation set of additional perturbations of varying severity.
Reliability diagrams and ECE values are calculated siimlar to \citet{zhou2021amortized} by binning exampels sorted by confidence into 20 bins, then calculating the average L1 norm between the average confidence and accuracy within each bin.

\subsection{Mislabel Detection}
For both CIFAR-10 and CIFAR-100 datasets we use a ResNet-18 model trained with the standard training procedure detailed in the section above, with early stopping calculated on a clean validation set.
We retrieve human and data-dependent noise directly from the relevant repositories, and generate asymmetric and symmetric noise locally.
Asymmetric noise is applied to CIFAR-10 with a predefined label mapping, and for CIFAR-100 we perform a mapping to the next fine-grained class in the corresponding coarse model class.

For Trac-IN we take model checkpoints every 20 epochs for a total of 10 checkpoints, then compute gradient norms on the full model.
EL2N and GraNd are computed by training a model for epoch 20 with 10 different seeds, then calculating average error L2 norm and gradient norms across all models. 
Self-IF is calculated using the same EKFAC estimated for IF-COMP with full gradients.
IF-COMP is calculated at $\beta = 0.001$ for all CIFAR-10 experiments and $\beta=1$ for all CIFAR-100 experiments.
Note that we fit the EKFAC on the dataset with noised labels.

\subsection{OOD Detection}
All baseline results are provided in the OpenOOD benchmark \citep{yang2022openood, zhang2023openood}, as well as the pre-trained LeNet and ResNet-18 models used for MNIST and CIFAR-10/CIFAR-100 experiments.
We use a temperature of $\beta = 2$ for MNIST experiments, $\beta = 0.001$ for CIFAR-10 experiments, and $\beta = 1$ for CIFAR-100 experiments. 
Temperature values are found by tuning on a held out validation set of OOD examples that share no classes with the ones used for testing.
This validation set is the same one used to tune hyperparameters for all benchmarks in the OpenOOD and are provided as a standard procedure from the framework \citep{zhang2023openood}.
We hypothesize that the extremely high temperatures necessary for strong CIFAR-10 results are a result of the stable and flat optima found with Batchnorm which prevent learning arbitrary labels without moving far away in weight and function space.

\subsection{Timing Comparisons}
For all experiments we use a single A40 GPU and attempt to use the same chunk size for all \texttt{vmap} operations.
However, ACNML on CIFAR-100 runs out of memory, so we reduce the chunk size from 4 to 1, meaning we perform the \texttt{vmap} only over the label space.
We follow the set of best practices provided in \citet{zhou2021amortized} and perform 5 optimization steps.
As in other experiments, we do not use the spatially uncorrelated activations (SUA) assumption for MNIST models, making the EKFAC computations slightly more expensive.


\end{document}

\end{document}